%% file: main.tex
\documentclass[10pt,journal,compsoc]{IEEEtran}

\usepackage{algorithm}
\usepackage{algorithmicx}

\usepackage{algpseudocode}
\usepackage{amsmath}
\usepackage{diagbox}
\usepackage{graphics}
\usepackage{epsfig}
\usepackage{threeparttable}
\usepackage{xspace}

\usepackage{graphicx}
\usepackage{amssymb}
\usepackage{threeparttable}

\usepackage{color}
\usepackage[normalem]{ulem}
\usepackage{multirow}
\usepackage{float}
\usepackage{amsfonts}
\usepackage{bm}
\usepackage{array}
\usepackage[table]{xcolor}
\usepackage{colortbl}
\usepackage{diagbox}
\usepackage{CJKutf8}
\usepackage{rotating}
\usepackage{booktabs}
\usepackage{overpic}
\usepackage{makecell} 

\usepackage{contour}
\usepackage{enumitem}
\usepackage{cite}
\usepackage{xcolor}

\usepackage[pagebackref=false,breaklinks=true,colorlinks,bookmarks=false]{hyperref}

\definecolor{mygray}{gray}{.92}

\makeatletter
\newcommand{\thickhline}{%
 \noalign {\ifnum 0=`}\fi \hrule height 1pt
 \futurelet \reserved@a \@xhline
}
\makeatother
\newcommand{\pub}[1]{\color{gray}{\tiny{[{#1}]}}}

\makeatletter
\DeclareRobustCommand\onedot{\futurelet\@let@token\@onedot}
\def\@onedot{\ifx\@let@token.\else.\null\fi\xspace}

\def\etal{\emph{et al}\onedot}

\makeatother

\begin{document}
\title{A Survey on fMRI-based Brain Decoding for Reconstructing Multimodal Stimuli}

\author{Pengyu Liu, Guohua Dong*, Dan Guo, Kun Li, Fengling Li, Xun Yang, Meng Wang, Xiaomin Ying*

\IEEEcompsocitemizethanks{
\IEEEcompsocthanksitem P. Liu is with the Center for Computational Biology, Beijing Institute of Basic Medical Sciences, Beijing, China, and the School of Computer Science and Information Engineering, Hefei University of Technology (HFUT), Hefei, 230601, China. (Email: lpynow@gmail.com)
\IEEEcompsocthanksitem G. Dong and X. Ying are with Center for Computational Biology, Beijing Institute of Basic Medical Sciences, Beijing, 100850, China. (Email: dgh1991.learn@gmail.com, yingxmbio@foxmail.com)
\IEEEcompsocthanksitem D. Guo and M. Wang are with the Key Laboratory of Knowledge Engineering with Big Data (HFUT), Ministry of Education, and the School of Computer Science and Information Engineering, Hefei University of Technology (HFUT), Hefei, 230601, China, and also with Institute of Artificial Intelligence, Hefei Comprehensive National Science Center, Hefei, 230026, China. (Email: guodan@hfut.edu.cn, eric.mengwang@gmail.com)
\IEEEcompsocthanksitem K. Li is with ReLER, CCAI, Zhejiang University, Hangzhou, 310058, China (Email: kunli.cs@zju.edu.cn). 
\IEEEcompsocthanksitem F. Li is with Australian Artificial Intelligence Institute, Faculty of Engineering and Information Technology, University of Technology Sydney, Ultimo, NSW 2007, Australia (Email: fenglingli2023@gmail.com)
\IEEEcompsocthanksitem X. Yang is with the School of Information Science and Technology, University of Science and Technology of China, Hefei, 230026, China. (Email: xyang21@ustc.edu.cn).

\IEEEcompsocthanksitem Corresponding author: (\textit{G. Dong and X. Ying})

}
}

\markboth{IEEE TRANSACTIONS ON PATTERN ANALYSIS AND MACHINE INTELLIGENCE}%
{Shell \MakeLowercase{\textit{et al.}}: Bare Demo of IEEEtran.cls for Journals}

\IEEEtitleabstractindextext{
\begin{abstract}
In our daily lives, we are exposed to a vast array of external stimuli, including images, sounds, and videos. As research into multimodal stimuli and neuroscience continues to advance, fMRI-based brain decoding has emerged as a powerful tool for understanding how the brain processes external stimuli and for exploring the mechanisms of human perception. The relationship between the brain and external stimuli is far from a simple linear correspondence; it involves intricate cognitive processes and individual cognitive differences. Therefore, decoding brain signals to reconstruct external stimuli not only helps uncover the complex mechanisms of brain perception but also offers new insights for the development of artificial intelligence, as well as advancements in disease treatment and brain-computer interface applications. In recent years, significant progress has been made in fMRI-based brain decoding, driven by advances in neuroimaging technology and image generation models. fMRI provides high-resolution spatial information to capture brain activity, enabling precise brain region signal acquisition for decoding. However, its low temporal resolution and inherent signal noise present challenges for decoding models. Concurrently, developments in image generation techniques, such as Generative Adversarial Networks (GANs), Variational Autoencoders (VAEs), and Diffusion Models (DMs), have significantly improved the quality of reconstructed images. Moreover, multimodal pre-trained models have further propelled the advancement of cross-modal decoding tasks. This survey systematically reviews recent progress in fMRI-based brain decoding, with a focus on studies that reconstruct stimulus content from brain signals generated by passive stimuli. It provides a detailed summary of the datasets and relevant brain regions employed in brain decoding tasks and categorizes and compares existing methods based on their model structures. Additionally, the survey qualitatively and quantitatively evaluates the performance of current mainstream models, offering an in-depth discussion of their effectiveness in this task. Finally, the survey highlights the major challenges in the field and proposes future research directions from both algorithmic and practical application perspectives, aiming to provide valuable guidance for the fMRI-based brain decoding domain. For more information and resources related to this survey, visit \url{https://github.com/LpyNow/BrainDecodingImage}.
\end{abstract}

}
\maketitle

\IEEEdisplaynontitleabstractindextext
\IEEEpeerreviewmaketitle

\begin{figure*}[t!]
\centering
\includegraphics[width=1\textwidth]{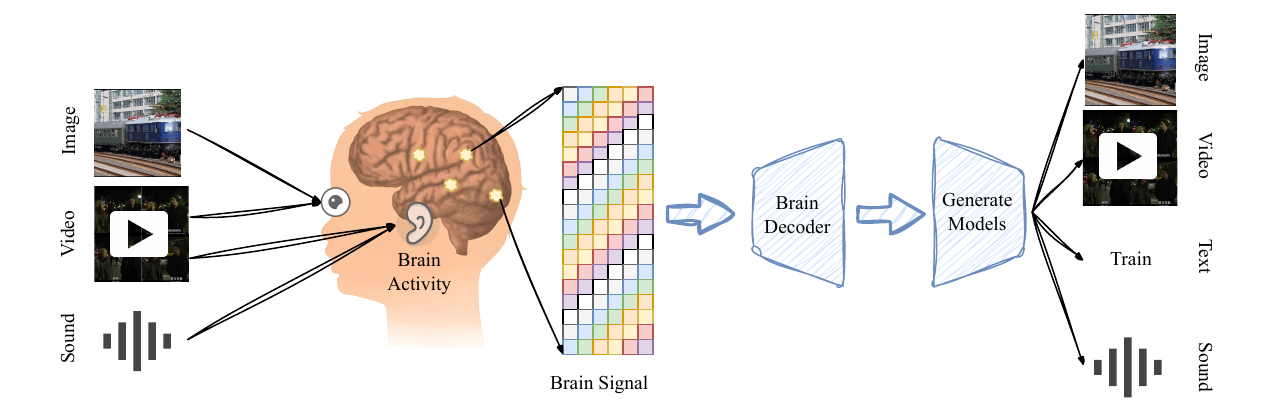}
\caption{The basic structure of fMRI-based brain decoding tasks: After the human body receives external stimuli, the brain generates signals in specific regions. By decoding these brain signals and using generative models for reconstruction, the stimulus signals received by the brain are restored.}
\label{fig:overview}
\end{figure*}

\section{Introduction}
\label{sec:introduction}

\IEEEPARstart{O}{ur} daily lives are filled with a multitude of multimodal stimuli~\cite{yeung2024neural, xia2018linked}. Various types of stimuli (visual, auditory, olfactory, etc.) are reflected in distinct signals within the brain~\cite{kasteleijn2004visual}. Decoding brain signals~\cite{haynes2009decoding,tong2012decoding,laconte2011decoding,saeidi2021neural} is essential for understanding how the human brain responds to external stimuli and for exploring the mechanisms of human perception. The relationship between external stimuli and brain signals is not a simple one-to-one correspondence~\cite{de2018decoding}. The brain’s response to external stimuli is not merely a passive process that generates corresponding signals; rather, it involves a complex cognitive process, which is also influenced by individual differences in perception. Therefore, reconstructing external stimuli from brain signals~\cite{lin2022mind, yang2024neurobind, liu2023decoding} can help us explore the intricate mechanisms by which the brain perceives external information. 

Research on brain decoding offers valuable insights~\cite{fan2020brain, chen2020topics} for the development of more advanced artificial intelligence. Brain decoding can contribute to building models that are more adept at understanding and generating human-like dialogue~\cite{huang2024brain}, thereby enhancing current large language models. Moreover, brain decoding can help us better understand ourselves and others, potentially altering underlying emotions and cognition. This research could also open new avenues for treating certain diseases. For instance, Alzheimer's disease~\cite{waldemar2007recommendations}, a neurodegenerative disorder potentially~\cite{hardy1999pathways} caused by neural degradation in the brain, might benefit from insights into how the brain encodes and decodes information, helping patients recover and restore memory. Additionally, brain decoding plays a crucial role in advancing brain-computer interfaces (BCIs)~\cite{xu2021review}, enabling humans to more easily control external devices. This would enhance human adaptability to the environment and broaden the range of human interaction with the world~\cite{lance2012brain, hu2022detecting}.

\subsection{Technological Advancements}\label{sec:technologicaladvancements}
With the development and advancement of time and related technologies, the cutting-edge technical requirements for brain decoding have gradually been met, providing crucial technical support for brain decoding research. These continuous innovations and improvements have enabled researchers to capture brain activity signals with unprecedented precision and decode high-quality brain information. Through high-resolution brain signal acquisition and analysis, a deeper understanding of the brain's response models across different tasks and scenarios can be achieved.

The development of neuroimaging technologies has provided significant support for research in brain decoding. These advancements enable the capture of clearer brain signals, thereby facilitating the decoding of high-quality brain information. Among these technologies, functional magnetic resonance imaging (fMRI)~\cite{heeger2002does}, electroencephalography (EEG)~\cite{teplan2002fundamentals}, magnetoencephalography (MEG)~\cite{da2013eeg}, and near-infrared spectroscopy (NIRS)~\cite{villringer1993near, cui2011quantitative} are several common neuroimaging techniques. Each of these methods, due to their characteristics, offers distinct advantages and contributions to brain decoding applications. 

Specifically, fMRI is a technique that indirectly detects brain signals by measuring blood oxygen level-dependent (BOLD) signals. When a specific region of the brain processes signals, the blood flow in that area increases, thereby enhancing the BOLD signal~\cite{logothetis2004interpreting}. fMRI captures these changes to provide high-resolution images of brain activity~\cite{kriegeskorte2007analyzing, laconte2011decoding}, offering researchers precise depictions of brain region activation. However, fMRI has a low temporal resolution, making it difficult to capture instantaneous changes in brain activity~\cite{christopher2008applications}, which limits the model's ability to quickly decode brain activity. 

EEG, on the other hand, directly records the electrical activity of the cortical brain using an array of electrodes, allowing it to reflect the synchronous firing processes of neurons in real time. However, EEG signals primarily originate from the brain's surface, resulting in low spatial resolution~\cite{michel2012towards}, which makes it challenging to locate deep brain activity. Additionally, EEG is highly susceptible to external signals and noise, necessitating complex data preprocessing and signal separation. 

MEG captures brain activity by recording the magnetic fields generated by neuronal activity. Meanwhile, NIRS~\cite{villringer1993near, strangman2002non} is a novel technique that measures blood oxygen changes in the cortical area using near-infrared light. Due to its portability and low-cost advantages, NIRS is more suitable for mobile environments or clinical applications.

In our survey, we primarily focus on fMRI-based datasets, model methodologies, and related approaches. This is mainly because fMRI offers high spatial resolution and provides a more precise stimulus-response correspondence, which aids researchers in better decoding the brain's response patterns across different tasks or scenarios. This is crucial for achieving high-quality outputs in brain decoding. Furthermore, most of the current research is based on fMRI signals, which is why we concentrated the focus of this survey on fMRI. However, we have also provided some discussion on research involving other brain signals.

In addition, the rapid development of generative models has significantly improved the quality and resolution of reconstructed signals, providing strong technical support for brain decoding tasks. Early efforts utilized Variational Autoencoders (VAEs)~\cite{doersch2016tutorial} to decode brain signals. VAEs integrate deep learning with probabilistic reasoning, enabling them to capture the underlying structure of data and generate a wide variety of images. 

Subsequently, Generative Adversarial Networks (GANs)~\cite{goodfellow2020generative} emerged as a powerful tool for image generation. GANs consist of two components: a generator and a discriminator. Through adversarial training, these models continuously refine their ability to generate realistic images~\cite{du2018reconstructing}, making GANs highly effective for image generation tasks. 

As research progressed, autoregressive models~\cite{terasvirta1994specification}, such as PixelCNN~\cite{van2016conditional} and PixelSNAIL~\cite{chen2018pixelsnail}, were introduced. These models generate images pixel by pixel, leveraging contextual information to model complex image structures and dependencies. This approach enables them to produce high-quality images with intricate details. 

The introduction of Diffusion Models (DMs)~\cite{song2020denoising} marked another significant milestone. DMs gradually add noise to real images and then use a reverse process to denoise and create new images. This technique has provided a crucial boost for reconstructing images of even higher quality and resolution.

Each stage of development in image generation models represents a significant advancement. These models not only provide strong support for brain decoding tasks but also offer substantial assistance in fields such as medical image analysis and artistic design.

At the same time, the introduction of large-scale pre-trained models, through joint training on multimodal data, has established a close connection between stimulus sources and natural language, opening up new research directions for cross-modal brain decoding tasks, particularly with OpenAI's CLIP~\cite{radford2021learning} model. CLIP achieves a close association between images and natural language by jointly training on both image and text data, providing valuable assistance for cross-modal tasks like brain decoding. Traditional computer vision models rely on large-scale labeled data for training, which requires significant time and computational resources and is often restricted to specific task scenarios. In contrast, CLIP uses unsupervised multimodal training on vast amounts of images and their corresponding natural language descriptions, making it adaptable to a wide range of multimodal tasks. Figure~\ref{fig:overview} illustrates the basic model design workflow for fMRI-based brain decoding tasks.

The advancements in these technologies, through joint training on multimodal data, have established a close connection between stimulus sources and natural language, providing new research directions for cross-modal brain decoding tasks. The synergistic development of these technologies not only enhances the accuracy and efficiency of brain decoding but also opens up vast possibilities for future breakthroughs in complex brain function analysis and cross-domain applications.

\subsection{Scope of the Survey}

Research on fMRI-based brain decoding has resulted in hundreds of related publications. Consequently, it is neither necessary nor feasible to include all such works in this survey. Instead, we have selected influential papers published in renowned journals and conferences, aiming to capture the most impactful contributions to the field. This survey primarily focuses on research from the past three years, reflecting the most recent advancements. However, to ensure the completeness and coherence of the discussion, we have also included seminal works and earlier studies that are relevant to the field or foundational to its current progress. Due to the limitations of space, knowledge, and scope, we acknowledge that some important studies might not have been included in this survey. We extend our apologies to the authors of these works. This survey seeks to provide a representative and structured overview rather than an exhaustive review of the entire literature.

\input{tabels/reviewsandsurveys}
\subsection{Realted Previous Reviews and Surveys}\label{sec:realted reviews and surveys}

In table~\ref{tab:reviewsandsurveys}, we also list previous reviews and surveys relevant to this investigation. fMRI-based brain decoding is a relatively novel task. Therefore, we revisit existing surveys related to this field. Among them, Du~\etal~\cite{du2019brain}. reviewed solutions proposed before 2019, focusing on early unified encoder-decoder frameworks and summarizing the positive impacts of deep generative models on brain encoding and decoding. Cao~\etal~\cite{cao2021computational}. concentrated on studies using human language as a stimulus source. Oota~\etal~\cite{oota2023deep}. specifically examined fMRI-based brain encoding and decoding architectures, providing an overview of related datasets and fundamental model structures. Mai~\etal~\cite{mai2023brain}. reviewed brain decoding models up to 2023, addressing models that reconstruct stimuli from various types of brain signals. Zhao~\etal~\cite{zhao2025diffusionmodelscomputationalneuroimaging} suggest that computational neuroimaging supports research on human cognition and behavior by analyzing brain images or signals, and the development of diffusion models also provides technical support for such research. They review the advancements in the application of diffusion models to neuroimaging tasks.

In recent years, with the rapid advancement of computational technologies, fMRI-based brain decoding research has made significant breakthroughs, fostering a deeper integration of visual cognition and neural decoding techniques. While the aforementioned review provides an overview of the progress in this field, it still has certain limitations, such as restricted content coverage, insufficient breadth, and a lack of comprehensive future outlooks. Compared to the existing literature, our survey offers the following key advantages:

\begin{enumerate}
    \item \textbf{Focus on Recent Advances:} We specifically highlight high-impact and high-quality research works that have emerged in recent years, systematically summarizing the latest technological progress in fMRI tasks. This approach avoids the information lag often seen in earlier surveys and ensures comprehensive coverage of emerging technologies.
    
    \item \textbf{Emphasis on fMRI Task-Oriented Model Design:} Unlike broader surveys that provide a general overview of brain signal decoding methods, our study focuses explicitly on model architecture design for fMRI tasks. We conduct an in-depth analysis of the core concepts, technical innovations, and application scenarios of different methodologies, offering clear and comprehensive guidance for future research.
    
    \item \textbf{Comprehensive Review and Comparative Analysis:} Our survey not only consolidates key aspects such as dataset characteristics, model structures, and actual decoding performance but also provides fine-grained analysis and in-depth discussions. This fills the gaps left by existing reviews in these crucial areas.
    
    \item \textbf{First-of-Its-Kind In-Depth Exploration:} To the best of our knowledge, our survey offers a more detailed and thorough discussion of critical aspects such as the strengths and limitations of different models and the challenges of cross-modal decoding—issues that have often been overlooked in prior reviews.
    
    \item \textbf{Extensive Future Outlook:} Our survey integrates current technological trends to provide a forward-looking perspective on the evolution of models and techniques, offering valuable insights into future directions in fMRI-based brain decoding research.
\end{enumerate}

\subsection{Article Structure}\label{sec:article structure}

\begin{figure*}[t!]
\centering
\includegraphics[width=1\textwidth]{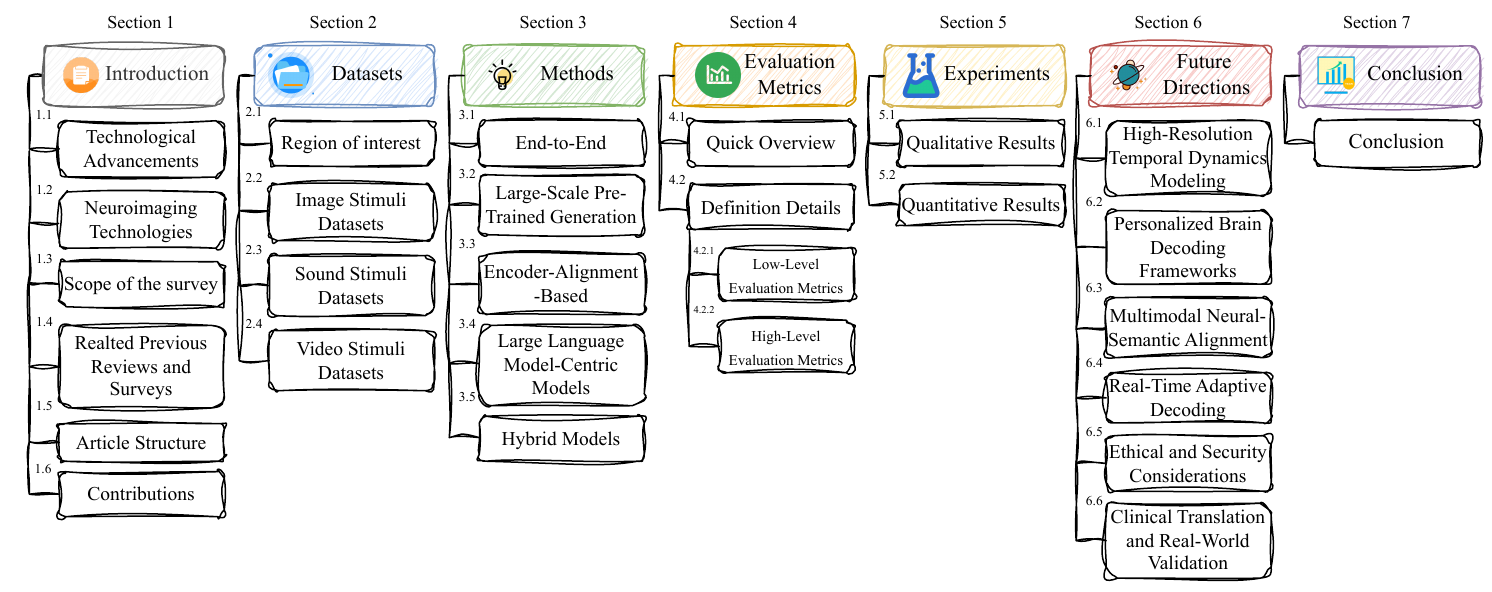}
\caption{A content overview covered in the survey.}
\label{fig:articlestructure}
\end{figure*}

In this survey, we systematically review and summarize the latest advancements in fMRI-based brain decoding tasks, as well as the implementation of various algorithms across different subtasks. Figure~\ref{fig:articlestructure} provides an overview of the content of this survey. In Section~\ref{sec:datasets}, we introduce commonly used regions of interest (ROIs) and relevant datasets for fMRI-based brain decoding. In Section~\ref{sec:methods}, we classify all types of algorithms based on their model structures. In Section~\ref{sec:evaluation metrics} describes the commonly used evaluation metrics. In Section~\ref{sec:experiments} presents the qualitative results and Quantitative Results of the models. In Section~\ref{sec:future} we also propose future research directions from the perspectives of algorithms and practical applications, offering insights that may contribute to the research and development of this field. Finally, in Section~\ref{sec:conclusion}, we conclude our survey and discuss the challenges of fMRI-based brain decoding tasks. 

\subsection{Contributions}

This survey comprehensively examines fMRI-based brain decoding, focusing on reconstructing stimuli from passively elicited brain signals. Specifically targeting generation tasks based on fMRI, the main contributions of this survey are as follows:  

\begin{enumerate}
\item Comprehensive Dataset Summary
\begin{itemize}
\item We systematically summarize a wide range of datasets related to brain decoding, with a primary focus on fMRI datasets.
\item To ensure completeness, we also include some related datasets outside the direct scope of fMRI-based decoding.
\item Additionally, we provide a unified discussion and organization of brain regions activated during external stimulation, linking these activations to their corresponding decoding tasks.
\end{itemize}
\item Categorization of Current Generation Models
\begin{itemize}
\item We review the development of state-of-the-art generation models.
\item The existing models and methodologies are systematically classified into categories, and the strengths and limitations of each category are critically analyzed.
\end{itemize}
\item Qualitative and Quantitative Analysis
\begin{itemize}
\item A comprehensive comparison of relevant models is presented, showcasing their qualitative and quantitative performance.
\end{itemize}
\item Insight into Current Progress and Future Directions
\begin{itemize}
\item We outline the current trends and challenges in the field, offering a forward-looking perspective on future research directions.
\item By consolidating insights from dataset organization and model design, we highlight potential avenues for advancing the field.
\end{itemize}
\end{enumerate}

Together, these contributions form a detailed, recent, and in-depth survey that not only consolidates the progress in the field but also distinguishes itself from prior reviews by offering novel insights and comprehensive coverage.  

\section{Datasets}
\label{sec:datasets}

Current datasets in the field of fMRI-based brain decoding are primarily collected in controlled experimental environments, using existing neuroimaging technologies to record brain signals when subjects passively receive external stimuli. These datasets typically include brain activity data, such as fMRI signals and EEG signals, as well as external stimuli signals used to evoke brain activity, usually in the form of images, text, video, and audio data.

When introducing and categorizing fMRI-based brain decoding datasets, we focus on brain signals passively generated in response to external stimuli. To systematically categorize and present these datasets, we classify them based on the type of stimulus, dividing them into three categories: image, sound, and video datasets. These three types of stimuli correspond to different brain activity patterns and decoding tasks. Therefore, this survey will categorize and introduce datasets according to these three types of stimuli.

Additionally, these various dataset types correspond to different brain decoding tasks. As shown in Figure 2, based on the types of stimuli and the brain signals decoded and reconstructed, we classify these tasks as follows: Image-to-Image (I2I), Image-to-Text (I2T), Speech-to-Speech (S2S), Speech-to-Text (S2T), Video-to-Video (V2V), and Video-to-Text (V2T). For example, the I2I task involves decoding and reconstructing corresponding image stimuli from brain activity induced by image stimuli, with other tasks defined similarly.

In table~\ref{tabel1_dataset}, we categorize these datasets based on their stimulus sources and provide detailed information, including the dataset's classification (Modality), data type (Type), publication date, journal of the dataset paper (Year \& Publish), number and source of stimuli (Experimental Paradigm), number of subjects (Subject), regions of interest (ROI), and the brain decoding tasks (Brain Task) that can be conducted using each dataset.

\input{tabels/datasets.tex}

\subsection{Neuroimaging Technologies} \label{sec:neuroimagingtechnologies}

Functional Magnetic Resonance Imaging (fMRI), Electroencephalography (EEG), Magnetoencephalography (MEG), and Near-Infrared Spectroscopy (NIRS) are several commonly used neuroimaging technologies.

\textbf{EEG:} EEG records the electrical activity of the brain cortex by placing electrodes on the scalp. It has high temporal resolution, capable of capturing millisecond-level neural activity changes. However, EEG has relatively low spatial resolution, making it difficult to pinpoint the exact cortical sources of brain activity. EEG signals are widely used in disease detection and sleep monitoring.

\textbf{MEG:} MEG measures the magnetic fields produced by neuronal electrical activity, using high-sensitivity devices such as Superconducting Quantum Interference Devices (SQUID). MEG offers excellent temporal resolution and good spatial resolution, enabling precise localization of brain signal sources in the cortex. However, MEG equipment is expensive, and the method requires a controlled environment. It is primarily used for brain function localization and studying the dynamic changes in neural networks.

\textbf{NIRS:} NIRS is an optical imaging technique that monitors or analyzes local blood oxygen levels to map tissue functional states. It provides a safe and non-invasive method for evaluating and mapping functional information in tissues. The main advantages of NIRS are its safety and non-invasiveness, making it applicable to special populations such as infants and the elderly. However, its lower spatial resolution limits its ability to measure brain signals only from the cortical regions.

\textbf{fMRI:} fMRI is a non-invasive neuroimaging technology widely used in brain function research and brain decoding fields. It primarily reflects neural activity through blood-oxygen-level-dependent (BOLD) signals. fMRI signal acquisition relies on magnetic resonance imaging (MRI) technology. During fMRI scanning, subjects usually perform specific tasks or remain at rest to activate different regions of the brain. The MRI machine uses powerful magnetic fields and radiofrequency pulses to acquire image data from various brain regions. These data are processed and analyzed to reveal the spatial and temporal characteristics of brain activity. Since its advent in the early 1990s, fMRI has experienced rapid development and evolution. Initially, fMRI was used to study the basic functional areas of the brain. With technological advancements, fMRI has been applied to more complex cognitive and emotional studies, such as language processing, memory, and emotion regulation. Additionally, combining fMRI with other neuroimaging techniques has further enhanced the understanding of brain function. Recent advancements in fMRI technology can be categorized into three areas: 1) Significant improvement in spatial resolution through the use of high-field MRI (e.g., 7T MRI) and improved imaging sequences, allowing for the observation of small brain structures; 2) Although fMRI's temporal resolution is relatively low, optimizing scan parameters and data processing methods enables more precise capture of dynamic brain activity changes; 3) Combining fMRI with other imaging techniques, such as EEG+fMRI, allows for simultaneous acquisition of both brain electrical activity and current changes, providing richer data for studying brain function. As the technology continues to improve, fMRI is widely applied in clinical diagnostics, brain-computer interfaces, and cognitive science research. Its non-invasive nature and high spatial resolution make it increasingly important in brain function research and clinical applications. Therefore, this survey primarily focuses on the field of fMRI-based brain decoding, while also showcasing related research using other brain signals.

\subsection{Region of interest}\label{sec:ROI}

Not all brain regions respond to external stimuli~\cite{namazi2015fractional}; only certain specific regions react to particular types of stimuli. These brain regions that respond to specific stimuli are referred to as Regions of Interest (ROIs). To facilitate the subsequent introduction of different types of datasets, in this section, we categorize and introduce these ROIs based on their response to different types of stimuli.

To facilitate the introduction of datasets with different types of stimulus sources, we have categorized and introduced the regions of interest (ROI).

\textbf{Early Visual Cortex:} The early visual cortex (EVC)~\cite{kosslyn2003early} is primarily responsible for processing basic visual information from the retina. These regions mainly handle fundamental visual features such as color, contrast, and edges. The V1 (Primary Visual Cortex)~\cite{tong2003primary} region is the first area in the brain that receives and processes visual information, dealing primarily with basic visual features like brightness, orientation, and contrast. The V2 (Secondary Visual Cortex)~\cite{rosa1999evolution} region, adjacent to V1, is the next step in visual information processing. It can handle slightly more complex visual features, such as the texture and depth of objects, helping the brain better understand the edges and shapes of objects. The V3 (Tertiary Visual Cortex)~\cite{shipp1995retinotopic} is mainly involved in shape and spatial arrangement of objects and is linked to depth and motion perception, enabling it to process information related to object position and shape changes. V4~\cite{roe2012toward} primarily handles color and shape perception, playing a crucial role in color processing and contour recognition, which is key in object recognition. V3A and V3B~\cite{press2001visual}: These two regions extend the functions of the V3 area, further involving motion and depth perception, with V3A being particularly important in perceiving 3D shapes and spatial layout.

\textbf{Higher Visual Cortex:} The higher visual cortex~\cite{yamins2014performance} is located at the top of the visual processing system and is mainly responsible for deeper analysis and processing of information that has already undergone preliminary processing. The Lateral Occipital Complex (LOC)~\cite{grill2001lateral} primarily handles object recognition, particularly processing the shape, size, and structure of objects. LOC is highly sensitive to the holistic perception of visual objects and is a key region in object decoding tasks. The Fusiform Face Area (FFA)~\cite{kanwisher2006fusiform} specializes in face recognition, located in the lower part of the temporal lobe. FFA is the specific area of the brain responsible for facial perception, crucial for recognizing facial features, and is often used for reconstructing facial images. The Parahippocampal Place Area (PPA)~\cite{epstein1999parahippocampal} is mainly responsible for scene recognition, especially for perceiving environmental and spatial layouts. It plays a critical role in distinguishing indoor and outdoor environments and is important in scene reconstruction tasks. The Retrosplenial Cortex (RSC)~\cite{vann2009does} is involved in spatial navigation and scene memory, crucial for long-term spatial memory and location perception. The Occipital Place Area (OPA)~\cite{dilks2013occipital} primarily processes information related to the spatial layout of visual scenes, especially in navigation tasks. It is mainly responsible for perceiving objects and spatial relationships in visual space. The Temporal Lobe (TL)~\cite{squire2004medial} is primarily responsible for auditory processing, language comprehension, and memory functions. Additionally, it plays a role in emotional responses and visual processing. The Occipital Lobe (OL)~\cite{sveinbjornsdottir1993parietal} is primarily responsible for visual processing, including the decoding of visual signals and the perception of spatial vision.

\textbf{Motion Visual Areas:} Motion visual areas~\cite{dupont1994many} are primarily responsible for processing motion and detecting and integrating moving objects within the visual field. The Middle Temporal Area (MT, V5)~\cite{van1981middle} is a key region for motion processing, responsible for detecting and processing moving objects in the visual field. MT is crucial for motion perception, and when external stimuli involve movement, MT responds to visual flow, speed, and direction of motion. The Lateral Intraparietal Area (LIP)~\cite{binkofski1998human} is associated with visual attention and eye movement control, responsible for integrating spatial information and the location of visual targets.

\textbf{Auditory-Related Areas:} Auditory-related areas~\cite{fullerton2007architectonic} are mainly responsible for processing basic auditory information, classification, and recognition tasks. The Primary Auditory Cortex (A1)~\cite{roe1990map} processes the basic features of sound. The Lateral Belt Area (LBelt)~\cite{schonwiesner2005hemispheric} and Parabelt Area (PBelt)~\cite{hackett2014feedforward} belong to the secondary auditory cortex, involved in the processing of complex auditory information such as sound recognition and localization. The Early Auditory Cortex (EAC)~\cite{huang2019associations} and Anterior Auditory Cortex (AAC)~\cite{brugge1985auditory} handle early and advanced auditory information, participating in complex tasks such as sound classification and recognition. A4 and A5~\cite{rolls2023auditory} are higher-level auditory processing regions involved in more advanced sound information analysis. The Retroinsular Cortex (RI)~\cite{benarroch2019insular} is related to spatial sound perception and can process the spatial localization of sound.

\textbf{Multimodal Perception Areas:} The Temporoparietal Occipital Junction (TPOJ)~\cite{igelstrom2017inferior} is a multisensory integration region, responsible for the combined processing of visual, auditory, and tactile information. TPOJ plays a crucial role in cross-modal tasks in brain decoding. The Premotor Cortex (PMC)~\cite{rizzolatti1996premotor} is mainly associated with action planning and execution, and it plays a role in multimodal perception and motor coordination.

\textbf{Face Recognition-Related Areas:} The Occipital Face Area (OFA)~\cite{pitcher2011role} handles the low-level processing of facial features, particularly the initial recognition of facial characteristics. Along with the FFA, OFA provides fundamental information during the decoding of facial information.

\textbf{Language and Motion-Related Areas:} Broca’s area~\cite{flinker2015redefining} is the region of the brain responsible for language generation and processing, located in the left hemisphere. The Superior Premotor Ventral Area (sPMv)~\cite{binkofski2006role} is related to motor planning, action generation, and language.

\textbf{Other Visual and Sensory Areas:} The Temporoparietal Junction (TPJ)~\cite{krall2015role} is involved in social cognition, attention control, and multisensory integration, particularly in the fusion of visual and auditory information. The Parietal Eye Fields (PEF)~\cite{brotchie2003head} are responsible for controlling eye movements and visual attention. The Parietal Lobe (PL)~\cite{fogassi2005parietal} is primarily responsible for the integration of sensory information, spatial perception, and motor coordination. The Frontal Lobe (FL)~\cite{fuster2002frontal} is closely associated with various higher cognitive functions.

\begin{figure}[t!]
\centering
\includegraphics[width=0.5\textwidth]{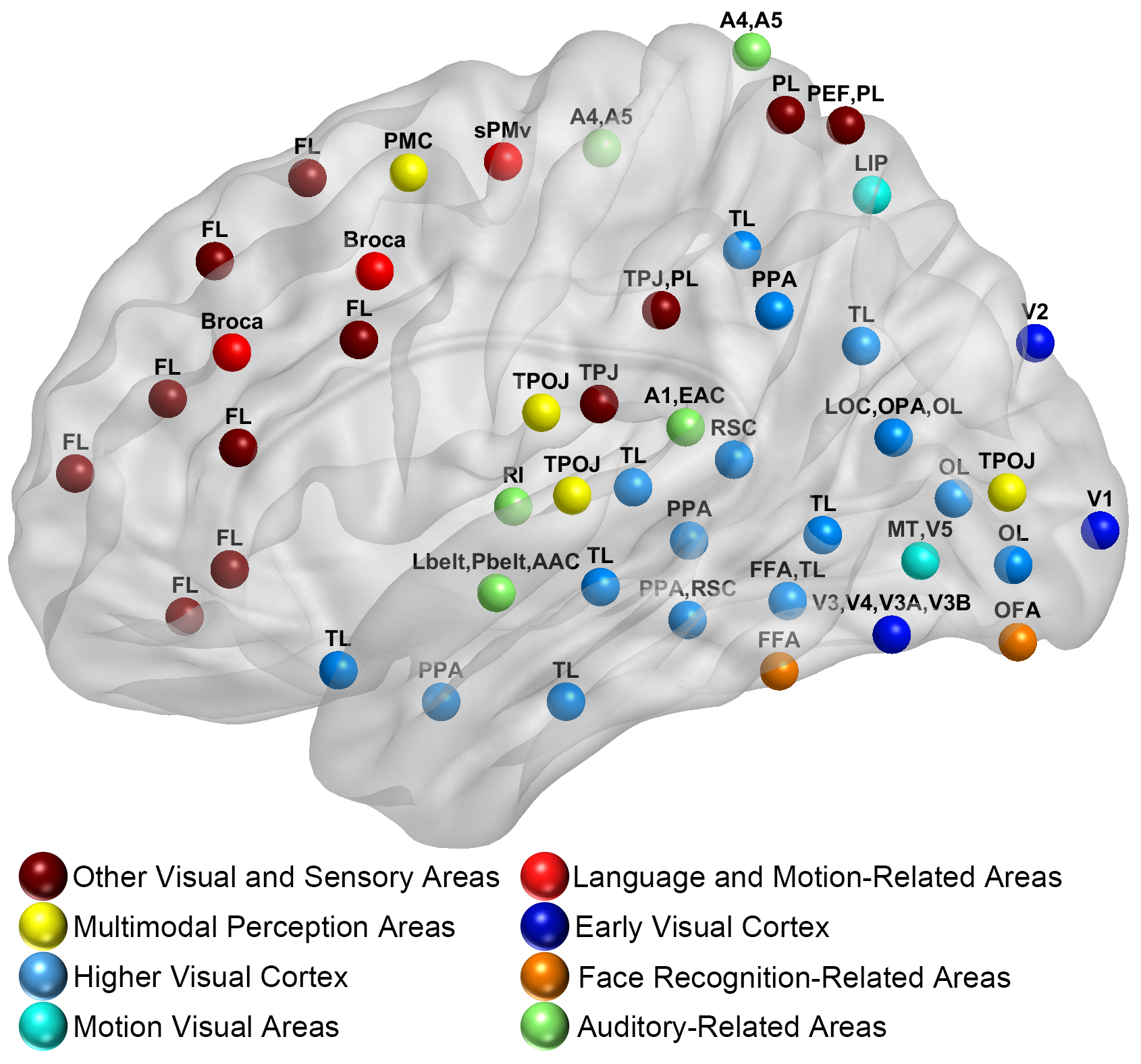}
\caption{We have illustrated the relative positional regions of the 38 subcategories of brain ROIs related to brain decoding, which are mentioned in the Section~\ref{sec:ROI}. Since ROI regions are irregular polygons and different ROIs may overlap to some extent, we use spheres to visually represent the approximate spatial relationships of the ROIs. The deep red represents the Other Visual and Sensory Areas, red represents the Language and Motion-Related Areas, yellow represents the Multimodal Perception Areas, deep blue represents the Early Visual Cortex, blue represents the Higher Visual Cortex, orange represents the Face Recognition-Related Areas, light blue represents the Motion Visual Areas, and green represents the Auditory-Related Areas. These ROIs are crucial for in-depth research on the stimulus-response relationship in the brain.}
\label{fig:ROI}
\end{figure}

The relative relationships between these regions are illustrated in Figure ~\ref{fig:ROI}.

\subsection{Image Stimuli Datasets}\label{sec:image stimuli datasets}

Early image datasets typically used simple shapes or patterns as stimuli, such as handwritten digits or characters. As computational power and hardware capabilities improved, stimulus sources became increasingly complex, incorporating more sophisticated images like facial images, grayscale natural images, and full-color natural scenes. In this section, we will provide a detailed overview of all datasets using image-based stimuli as comprehensively as possible.

The Binary Contrast Patterns (BCP)~\cite{miyawaki2008visual} dataset is an early dataset that uses low-resolution composite images generated through binary contrast patterns. It provides fMRI information from the primary visual cortex (V1, V2, and V3) of four subjects, making it suitable for I2I reconstruction tasks. 

The Handwritten Digits (HWD)~\cite{van2010neural} dataset, published in 2010 in Neural Computation, includes 2,106 handwritten grayscale digit samples selected from MNIST~\cite{deng2012mnist}. It primarily explores brain activity during the digit recognition process and tests the fMRI information of one subject, supporting I2I tasks. 

The Handwritten Characters (BRAINS)~\cite{schoenmakers2013linear} dataset consists of 720 handwritten character samples selected from a handwritten character recognition dataset~\cite{van2009new} as stimuli. It aids in achieving I2I reconstruction tasks by testing the primary visual cortex (V1) of three subjects. 

The BCP, HWD, and BRAINS datasets are considered relatively early datasets. Limited by the technology and hardware conditions at the time, the stimuli used were simple characters or patterns, and the images had lower resolutions.

The BOLD5000 dataset~\cite{chang2019bold5000} is a large-scale fMRI neuroimaging dataset containing brain activity data for 4,916 images sourced from the SUN, COCO, and ImageNet datasets. It records the fMRI responses of 4 subjects as they viewed image stimuli, with four specific ROIs: PPA, RSC, OPA, and EVC. This dataset includes various everyday scenes and objects, making it applicable to both I2I and I2T tasks. 

The Natural Scene Dataset (NSD)~\cite{allen2021massive} is the largest and most widely used fMRI dataset to date, capturing brain activity data for 8 subjects in response to image stimuli. Currently, complete data collection has been achieved for subjects 1, 2, 5, and 7, making their data the primary focus of most research, and our survey also centers on these four subjects. The NSD uses images from the COCO dataset as stimuli, with each subject viewing 9,000 repeated images and 1,000 unique images, for a total of 70,566 presentations across extended testing sessions. This extensive setup provides stable and consistent experimental data. The NSD dataset includes several visual cortex ROIs related to visual processing and object recognition, primarily: V1, V2, V3, V4, LOC, FFA, and PPA. These data are widely applicable to both I2I and I2T tasks. 

The Generic Object Decoding (GOD)~\cite{horikawa2017generic}dataset is an fMRI-based dataset that records brain activity from five subjects as they viewed 1,250 images sourced from the ImageNet dataset. The GOD dataset also provides ROI data covering regions such as V1 through V4, LOC, FFA, and PPA, supporting both Image-to-Image (I2I) and Image-to-Text (I2T) decoding tasks. 

The Deep Image Reconstruction (DIR)~\cite{shen2019deep} dataset records the brain activity of three subjects while viewing 50 artificial shapes and letters. It identifies the ROIs as V1 to V4, LOC, FFA, and PPA. The Vim-1~\cite{kay2008identifying} dataset is an early fMRI dataset that contains a total of 1,870 samples, including 1,750 training samples and 120 test samples sourced from the Commercial Digital Library and the Berkeley Segmentation Dataset. Two subjects were tested, and the defined ROI regions in this dataset are V1 to V4. During data collection, the presentation time of images was strictly controlled to ensure high-quality data acquisition. 

The CelebrityFace~\cite{vanrullen2019reconstructing} dataset is relatively new, comprising 108 samples with 88 training samples and 20 test samples sourced from the CelebA dataset, which is a large facial image dataset featuring various celebrity faces. The rich detail of facial images compared to natural images demands a higher level of precision during the image reconstruction process. The CelebrityFace dataset provides data from four subjects concerning four facial processing-related ROIs: TL, OL, PL, and FL. The DIR, Vim-1, and CelebrityFace datasets are all applicable to I2I tasks. The Human Connectome Project (HCP)~\cite{van2013wu} dataset can be widely applied in neuroscience research, including the localization of brain functional areas, connectome analysis, and the investigation of neural activity patterns across different subtasks. 

The Multi-Scale Connectivity (MSC)~\cite{gordon2017precision} dataset is relatively smaller, providing fMRI information from 10 subjects and focusing on analyzing the multi-scale connectivity of the brain in small sample sizes. In recent studies, the MSC dataset from HCP has often been used for model pre-training. Additionally, the EEG-VOA~\cite{spampinato2017deep} dataset is based on EEG data, using 2,000 images from ImageNet as external visual stimuli. It offers high-resolution EEG signals with 128 channels from six subjects, allowing researchers to capture changes in attention as subjects process different visual objects. Although EEG datasets typically do not directly specify ROI regions, it is possible to infer visually related brain areas through the analysis of specific frequency bands and channels. By decoding EEG signals, researchers can reconstruct subjects' responses to visual stimuli, making it well-suited for I2I tasks.

\subsection{Sound Stimuli Datasets}\label{sec:sound stimuli datasets}

Sound stimulus datasets typically feature a diverse range of stimuli, including music, spoken text, or narratives. We categorize these datasets as sound stimulus-driven datasets. In this section, we summarize and describe such datasets.

The Brain Sound Recognition (BSR)~\cite{park2023sound} dataset is based on fMRI data, using 1,250 sound clips from the VGGSound dataset as stimuli. It provides fMRI information from multiple regions of interest (ROIs) across five subjects, including both primary and secondary auditory cortices such as A1, LBelt, Pbelt, A4, and A5. The BSR dataset is crucial for understanding how the brain processes sound, primarily applied to reconstruct the auditory stimuli faced by subjects through fMRI decoding, thus facilitating S2S tasks. 

The Narratives~\cite{nastase2021narratives} dataset involves 27 stories composed by researchers, narrated to 345 subjects while recording their brain fMRI activity. These stories encompass various types, aiming to capture brain activity patterns as subjects comprehend different story contents. The dataset includes ROIs such as A1, Mbelt, LBelt, and RI, which are closely related to language understanding. Given that the stimuli are custom-written story texts, the Narratives dataset is suitable for S2T tasks. 

The End-to-End Temporal Classification of Audio Signals (ETCAS)~\cite{guo2023end} dataset is based on EEG data, using 50 audio clips from the TIMIT dataset as stimuli. It tested 50 subjects, collecting high-resolution EEG signals via 24 channels. The ETCAS dataset is mainly adapted for S2S tasks. 

The MusicGener~\cite{nakai2022music} dataset comprises 540 music clips used as stimuli, tested on three subjects, with 480 clips designated as a training set and 60 as a test set, representing various musical styles. 

The MusicAffect~\cite{daly2020neural} dataset evaluated 21 subjects with synthesized or classical music fragments, recording their brain fMRI while listening. 

The GTZan~\cite{ferrante2024r} dataset selected 540 music clips across multiple genres as stimuli, analyzing brain activity by recording fMRI from five subjects. MusicGener, MusicAffect, and GTZan datasets are all applicable for S2S tasks. 

Finally, the Story~\cite{nastase2020leveraging} dataset uses ten stories as stimuli, recording fMRI information from 149 subjects during the process. It includes multiple ROIs such as EAC, AAC, TPOJ, and PMC, making it suitable for S2S tasks.

\subsection{Video Stimuli Datasets}\label{sec:video stimuli datasets}
Datasets utilizing video stimuli typically employ clips from films or videos as the stimulus source, necessitating continuous recording of brain activity. Furthermore, the complex temporal relationships between successive video frames pose significant challenges for the model in recognizing and reconstructing the signals, as each frame contains not only the main subject but also various contextual elements.

The Cam-CAN~\cite{taylor2017cambridge} dataset captures the brain activity of 656 participants while viewing movie clips, utilizing multiple fixed-length film segments as stimuli. 

The DNV~\cite{wen2018neural} dataset includes videos sourced from YouTube, comprising 374 clips for training and 598 for testing, totaling 972 segments. This dataset analyzes brain responses by capturing fMRI data from three participants, focusing on ROI areas such as V1, V2, V3, V4, LO, MT, FFA, PPA, LIP, TPJ, and PEF during video viewing. 

The VER~\cite{nishimoto2011reconstructing} dataset derives its stimuli from the Apple Quick-Time gallery and YouTube, featuring 12,600 video clips that encompass a variety of contexts and content. This dataset also records fMRI information from three participants, focusing on several ROIs, including V1, V2, V3, V3A, and V3B. 

The STNS~\cite{seeliger2019large} dataset utilizes 30 episodes from "Doctor Who" as stimuli, capturing the fMRI data of a single participant while watching the videos. The ROIs provided in this dataset include V1, V2, V3, MT, AC, FFA, LOC, and OFA. 

The CLSR~\cite{tang2023semantic} dataset is distinctive, as it combines silent video clips with narrated stories sourced from "The Moth Radio Hour" and "Modern Love." It records brain activity from three participants, focusing on ROIs such as AC, Broca, and sPMv. 

The NNDB~\cite{aliko2020naturalistic} dataset uses 10 full-length films as stimuli, with the individual clips being relatively long, capturing brain data from 86 participants. The extended duration of these films presents additional challenges for model design. All the aforementioned video datasets are applicable to V2V tasks, while the CLSR dataset can also be utilized for S2T tasks. 

Video-driven datasets inherently merge characteristics of both image and sound stimuli, requiring models to capture finer details of the stimulus source while also accounting for the logical information embedded within the temporal sequence, thereby facilitating reconstruction tasks.

\section{Methods}\label{sec:methods}

\begin{figure*}[t!]
\centering
\includegraphics[width=0.7\linewidth]{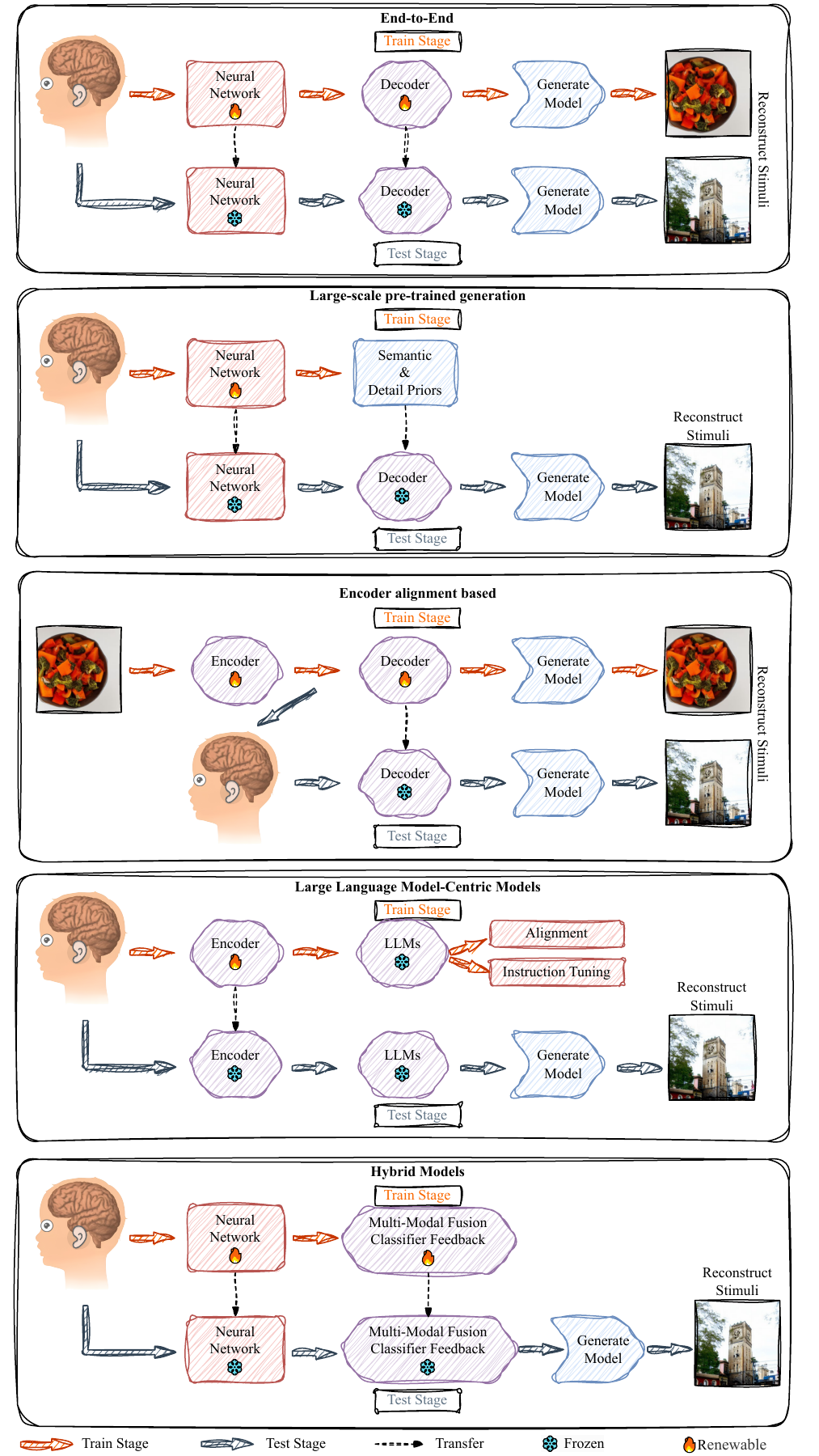}
\caption{Overview of Models for fMRI-Based Brain Decoding Tasks.}
\label{fig:Models}
\end{figure*}

In fMRI-based brain decoding tasks, the typical process involves using brain signals as input to capture multimodal information and reconstruct it into the original stimuli. Therefore, the usual model design consists of two main components: a network to capture multimodal input information and a generative model to reconstruct the stimulus. Research in brain decoding is limited by the noise in brain signals, inherent specificity across different subjects, as well as challenges in dataset size and the fusion of brain signals with external multimodal information. 

\begin{figure*}[t!]
\centering
\includegraphics[width=0.95\linewidth]{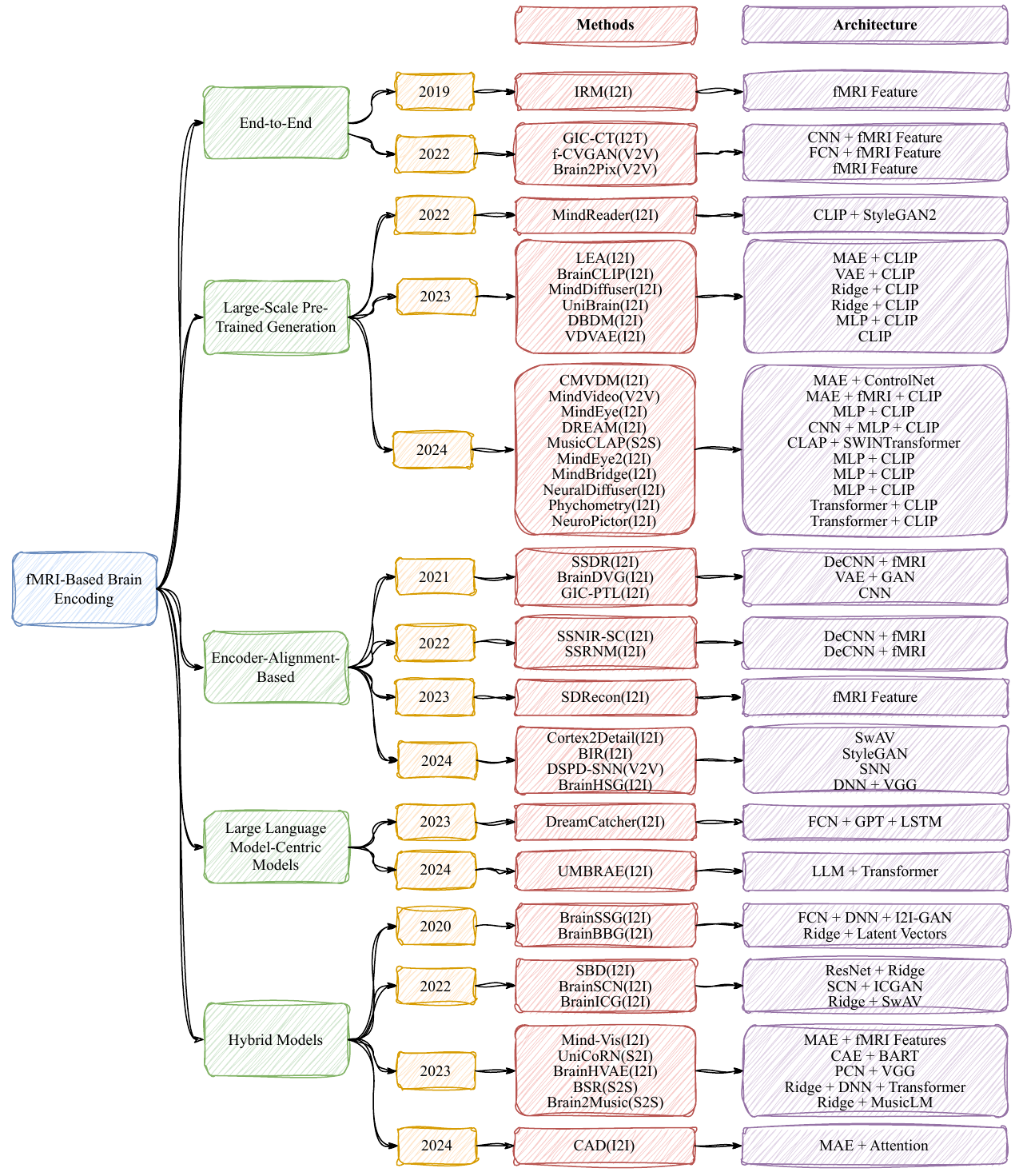}
\caption{Model Classifications for fMRI-Based Brain Decoding Tasks.}
\label{fig:modelclassifications}
\end{figure*}

In the following sections, we summarize the current fMRI-based brain decoding models and categorize them according to their primary model architectures. We divide these models into three main categories: end-to-end models, large-scale pretraining-based models, encoder-alignment-based models, Large Language Model-Centric Models, and Hybrid Models. In Figure~\ref{fig:Models}, we present schematic diagrams of the basic architectures of these models. Additionally, we present the classification of existing representative model algorithms in Table~\ref{tab:models} and Figure~\ref{fig:modelclassifications}.

\subsection{End-to-End}\label{sec:end to end}

End-to-End Models directly input fMRI signals into a network to achieve the reconstruction mapping from brain signals to target modality information~\cite{dale2001spatiotemporal, palazzo2020decoding, li2022multi}. The core idea of this approach is to treat the entire decoding process as an optimization problem, avoiding the need for manual feature extraction and alignment steps, which simplifies the process and reduces both manual intervention and the complexity of model design~\cite{cao2024survey}. This approach enables end-to-end models to fully leverage the strengths of deep learning in automatic feature learning and modeling complex nonlinear relationships, thereby capturing the deep semantic associations between fMRI signals and external stimuli~\cite{de2022neuroscout}. This direct mapping significantly enhances the model's adaptability and efficiency, making the reconstruction from brain signals to target stimuli more accurate and effective.

One key advantage of end-to-end models is their simplified structure, as the network design is directly focused on the target task without the need for complex intermediate feature extraction and alignment steps. This streamlines the training process, making it clearer and more efficient. Moreover, end-to-end architectures allow training and data processing to be performed within a unified framework, reducing system complexity and improving operational convenience. However, challenges arise due to the high dimensionality, low temporal resolution, and strong noise inherent in fMRI signals, which place high demands on the model's learning capability. For example, the complex and nonlinear relationship between noise in the fMRI signal and the true stimuli makes it difficult for end-to-end models to effectively extract useful semantic information from the noise during training, thus affecting the quality of the final reconstruction.

Furthermore, the high noise and blurriness of fMRI data pose significant challenges for end-to-end methods. Although this approach allows for direct mapping from signals to stimuli, the low resolution~\cite{kriegeskorte2007analyzing, liu2016noise, kundu2017multi, thomas2002noise, kirwan2007high} and high noise characteristics of fMRI signals often result in blurred or semantically meaningless outputs during stimulus reconstruction, preventing high-quality image reconstruction. Another issue is that existing fMRI datasets are typically small, and the high cost of data collection leads to a scarcity of high-quality paired data needed for training end-to-end models. In cases of insufficient data, the model's generalization ability and robustness are often compromised, reducing its effectiveness in real-world applications.

The "black-box" nature of deep learning models is also a major drawback of end-to-end approaches. In tasks that require model interpretability, the generated results and the model's internal decision-making process are often difficult to intuitively understand or explain. Therefore, ensuring model performance while improving interpretability is a significant challenge in the application of end-to-end models.

Despite these challenges, end-to-end methods still show great potential in brain decoding tasks. Through refined network design and optimization, these methods can effectively learn meaningful patterns from complex fMRI signals, achieving accurate mappings between brain signals and stimulus information. To further improve the performance of end-to-end methods, different model architectures can be optimized and adjusted, and various strategies can be applied based on the characteristics of specific tasks and datasets to address existing shortcomings.

Although End-to-End methods directly map fMRI data to stimuli~\cite{zhang2022cnn}, these methods can still be further optimized and subdivided according to different model architectures and optimization objectives. Below, we have summarized some End-to-End model categories and their primary optimization directions:

\begin{enumerate}
    \item Based on Convolutional Neural Networks (CNN):
    Convolutional Neural Networks (CNNs)~\cite{gu2018recent} are widely used in computer vision tasks and are well-suited for structured data. For fMRI signals, CNNs mainly extract local spatiotemporal features from brain signals through convolutional layers. For the input brain signal \( \mathbf{X} \in \mathbb{R}^{N \times D} \), where \( N \) is the number of time steps and \( D \) is the feature dimension at each time point, the goal of the End-to-End network is to learn a mapping \( \mathbf{X} \rightarrow \mathbf{Y} \), where \( \mathbf{Y} \) is the target stimulus information.

    CNNs transform the signal \( \mathbf{X} \) into higher-level feature representations \( \mathbf{X}_{conv} = \sigma(W * \mathbf{X} + b) \), where \( W \) is the convolution kernel, \( b \) is the bias term, and \( \sigma \) is the activation function. After multiple layers of convolution, pooling, and fully connected layers, the final output \( \hat{\mathbf{Y}} \) is a reconstruction of the target stimulus.
    
    For CNN-based End-to-End model design, the loss function is typically the Mean Squared Error (MSE)~\cite{eldar2004robust}:
    \begin{equation}
        L_{\text{CNN-E2E}} = \frac{1}{N} \sum_{i=1}^N \left\| \hat{\mathbf{Y}}_i - \mathbf{Y}_i \right\|^2
    \end{equation}
    
    where \( \hat{\mathbf{Y}}_i \) is the predicted value for the \( i \)-th sample, and \( \mathbf{Y}_i \) is the true value.

    \item Based on Recurrent Neural Networks (RNN):
    Recurrent Neural Networks (RNNs)~\cite{schuster1997bidirectional} are suitable for handling sequential data and capturing long-term dependencies in signals. For fMRI data, RNNs can model the temporal evolution of signals and capture dynamic features at each time step. For input signals \( \mathbf{X} = \{x_1, x_2, \dots, x_T\} \), where \( T \) is the number of time steps and each \( x_t \in \mathbb{R}^D \) is the fMRI feature at time step \( t \), the RNN computation is as follows:
    
    \begin{equation}
        h_t = f(W_h h_{t-1} + W_x x_t + b_h)
    \end{equation}
    
    where \( h_t \) is the hidden state, \( W_h \) and \( W_x \) are weight matrices, \( b_h \) is the bias term, and \( f \) is the activation function. The final output is:
    
    \begin{equation}
        \hat{\mathbf{Y}}_t = g(W_h h_t + b_y)
    \end{equation}
    
    where \( g \) is the activation function of the output layer.
    
    For RNN-based End-to-End model design, the loss function is typically the Mean Squared Error (MSE):
    
    \begin{equation}
        L_{\text{RNN-E2E}} = \frac{1}{T} \sum_{t=1}^T \left\| \hat{\mathbf{Y}}_t - \mathbf{Y}_t \right\|^2
    \end{equation}

    \item Based on Generative Adversarial Networks (GAN):
    Generative Adversarial Networks (GANs)~\cite{goodfellow2020generative} employ a generator and a discriminator for adversarial training, enabling the generation of images or signals that resemble the target stimuli. For fMRI decoding tasks, the generator \( G \) learns to reconstruct the target stimulus \( \mathbf{Y} \) from the fMRI signal \( \mathbf{X} \), and the discriminator \( D \) is used to distinguish between the generated data and real data. The generator's goal is to maximize the discriminator's error, with the optimization objective being:
    
    \begin{equation}
        L_G = - \mathbb{E}_{\mathbf{X} \sim p_{\mathbf{X}}} [\log D(G(\mathbf{X}))]
    \end{equation}
    
    The discriminator's goal is to maximize its ability to distinguish between real and generated samples, with the optimization objective being:
    
    \begin{equation}
        L_D = L_G - \mathbb{E}_{\mathbf{X} \sim p_G} [\log (1 - D(G(\mathbf{X})))]
    \end{equation}

    \item Based on Transformer:
    The core idea behind Transformer-based End-to-End methods is the self-attention mechanism~\cite{vaswani2017attention}, which can capture information from other positions in a sequence, making it well-suited for handling long-range dependencies in sequential data. In fMRI decoding, the input signal can be represented as a sequence \( \mathbf{X} = \{x_1, x_2, \dots, x_T\} \), where each \( x_t \in \mathbb{R}^D \) is the feature vector of the fMRI at time step \( t \). The self-attention mechanism of Transformer is computed as:
    
    \begin{equation}
        \text{Attention}(Q, K, V) = \text{softmax}\left( \frac{Q K^T}{\sqrt{d_k}} \right) V
    \end{equation}
    
    where \( Q \), \( K \), and \( V \) are query, key, and value matrices, respectively, and \( d_k \) is the dimension of the key. Through multiple layers of self-attention, the final output sequence can be used for stimulus reconstruction. The loss function is typically the Mean Squared Error (MSE), as shown below:
    
    \begin{equation}
        L_{\text{Transformer-E2E}} = \frac{1}{T} \sum_{t=1}^T \left\| \hat{\mathbf{Y}}_t - \mathbf{Y}_t \right\|^2
    \end{equation}
\end{enumerate}

Thus, while end-to-end models are straightforward and direct for exploring the correspondence between brain signals and stimuli, their training increases the demand for high-quality datasets. Additionally, the "black box" nature of deep learning limits the interpretability of these models.

\subsection{Large-Scale Pre-Trained Generation}\label{sec:large-scale pre-trained generation}

In modern brain decoding tasks, models that leverage large-scale pretraining datasets are gradually becoming the mainstream approach. These methods employ a staged training strategy, where the pretraining phase is conducted on external large-scale multimodal datasets (e.g., images, text, and audiovisual data), allowing the model to learn rich semantic information from vast amounts of unlabeled data. The key to this strategy is that the model, without explicit labels, can capture multimodal features and distribution patterns through the correlations between data, thereby enhancing its semantic modeling and generative capabilities~\cite{tong2012decoding}.

One of the leading pretrained models in this domain is CLIP~\cite{radford2021learning} (Contrastive Language-Image Pretraining), which uses multimodal learning to capture complex semantic relationships between images and text, providing strong support for decoding fMRI signals and external stimuli. CLIP employs a dual-tower architecture~\cite{yuan2024video}, with independent encoders designed for images and text. Through a contrastive learning strategy, CLIP optimizes the embedding spaces for text and images so that the embeddings of corresponding text and images are as close as possible, while embeddings from different text-image pairs are pushed apart. This design allows CLIP to learn cross-modal semantic relationships adaptively during training, enabling it to capture multimodal correlations through shared representations of images and text in downstream tasks~\cite{fu2013learning}.

The training process of CLIP uses a contrastive loss function, which optimizes the similarity between images and text to achieve cross-modal alignment. For a batch of images and text, CLIP maps them into their respective embedding spaces, yielding image embeddings \( v_i \) and text embeddings \( t_i \). The loss function is optimized as follows:

\begin{equation}
    L_{\text{Pretrain}} = -\frac{1}{N} \sum_{i=1}^{N} \log \frac{\exp(v_i \cdot t_i / \tau)}{\sum_{j=1}^{N} \exp(v_i \cdot t_j / \tau)}
\end{equation}

Where \( v_i \) and \( t_i \) represent the embeddings of the \( i \)-th image and text, \( \tau \) is the temperature parameter controlling the scaling of the embedding space, and \( N \) is the batch size. The goal of \( L_{\text{Pretrain}} \) is to maximize the similarity between the embeddings of each image-text pair while minimizing the similarity between them and other text or image embeddings. Through this method, CLIP effectively learns the shared semantic space between images and text. After pretraining, the model can be fine-tuned or feature-aligned for specific fMRI decoding tasks by transferring the prior knowledge learned during pretraining.

To adapt the pretrained model to a specific fMRI decoding task, fine-tuning or feature alignment is typically required to map between modalities~\cite{rastegarnia2023brain, sarafraz2024domain}. Fine-tuning usually involves modifying the last few layers of the original model and training it with specific fMRI data. The loss function for fine-tuning is typically task-dependent, such as the Mean Squared Error (MSE) for image reconstruction tasks or Cross-Entropy Loss for classification tasks.

When mapping between different modalities, feature alignment becomes a crucial step. For the features \( \chi_{\text{fMRI}} \) from fMRI signals and \( \chi_{\text{stimuli}} \) from the stimulus data, the goal is to map them into a shared space \( S \) through an alignment loss for effective decoding. The feature alignment loss is as follows:

\begin{equation}
    L_{\text{align}} = \| \chi_{\text{fMRI}} W_f - \chi_{\text{stimuli}} W_i \|_2^2
\end{equation}

Where \( W_f \) and \( W_i \) are the projection matrices for fMRI signals and stimulus data, and \( \| \cdot \|_2 \) is the L2 norm used to measure the distance between the features after mapping them into the shared space. Through this process, fMRI signals and target stimuli can be mapped into a common latent space, enabling effective cross-modal decoding.

In addition to CLIP, other pretrained models~\cite{wang2023pre} have also been applied to brain decoding tasks, offering various strategies to enhance decoding accuracy and semantic understanding. For example, the VLP (Vision-Language Pretraining)~\cite{kim2021vilt, dou2022empirical, xu2021e2e} model, ALIGN (A Large-scale Image and Noisy-Text)~\cite{jia2021scaling}, and UNITER (Universal Image-Text Representation)~\cite{chen2020uniter} have all achieved significant success in vision-language pretraining and provide robust support for cross-modal feature learning.

The VLP models, such as ViLBERT~\cite{lu2019vilbert} and VisualBERT~\cite{li2019visualbert}, combine image and text into a multimodal framework for learning and are capable of modeling both visual and textual information simultaneously. These models adopt a structure similar to BERT~\cite{devlin2019bert}, encoding both images and text to capture their relationships and offering a potential mapping framework for fMRI signal decoding tasks.

The ALIGN model, similar to CLIP, is also based on a contrastive learning framework, learning multimodal representations by aligning the embedding spaces of images and text. ALIGN leverages large datasets to optimize the relationship between image-text pairs, offering transfer learning capabilities for tasks such as brain decoding.

The UNITER model uses a joint Transformer architecture to process both image and text information, achieving good performance in vision-language tasks. By training on a joint representation, UNITER enables visual and textual information to be understood within the same space, which holds significant potential for modeling the relationship between fMRI signals and external stimulus information.

When using pretrained models for decoding, especially for image or video generation models, the computational complexity is often high, particularly when dealing with high-resolution images or video generation. These generation tasks typically involve multiple convolutional layers, fully connected layers, or Transformer layers, leading to a large number of parameters and computational demands. In high-resolution stimulus image or video reconstruction~\cite{chen2023seeing}, this can result in significant computational overhead. To address this challenge, strategies such as model pruning, quantization, and multi-scale generation are commonly employed to reduce computational burdens. For instance, model pruning reduces network size by removing unnecessary neurons or connections, while quantization lowers memory usage and computational complexity by converting floating-point numbers into lower-precision values, ensuring that the model maintains accuracy while having faster inference speeds.

Furthermore, the design of fine-tuning and training strategies is critical to the success of pretrained models in brain decoding tasks. Balancing the training strategies between the pretraining and fine-tuning stages~\cite{zhou2023dr}, selecting appropriate learning rates, optimizers, and task-specific loss functions are all important factors affecting the final decoding performance. Typically, the loss function during the fine-tuning phase is adjusted according to the target task, such as using Mean Squared Error (MSE) for reconstruction tasks:

\begin{equation}
    L_{\text{fine-tune}} = \frac{1}{N} \sum_{i=1}^{N} \| \hat{Y}_i - Y_i \|
\end{equation}

\include{tabels/model.tex}

For brain decoding model design based on large-scale pretrained models, methods using CLIP and other multimodal pretrained models can effectively leverage cross-modal learning strategies to support fMRI decoding tasks. These models, through the semantic information learned in the pretraining phase, help mitigate issues related to data sparsity and cross-subject robustness, and enhance model performance via cross-modal alignment. However, challenges remain in handling modal differences, optimizing computational resource usage, and designing suitable fine-tuning mechanisms. In the future, with ongoing improvements in model architecture and training strategies, brain decoding technologies based on large-scale pretraining models will exhibit tremendous potential across various fields.

\subsection{Encoder-Alignment-Based}\label{sec:encoder alignment based}

The design of encoder-based alignment models aims to map fMRI signals and external stimuli (such as images, text, audio, etc.) into a shared latent feature space, facilitating effective alignment and decoding across modalities. Unlike traditional brain decoding methods, these models directly extract meaningful spatiotemporal features from data through automatically learned encoders~\cite{han2019variational} and align the raw fMRI signals with external stimuli through nonlinear mappings. This approach allows for a more precise capture of the deep semantic relationships between different modalities. The core advantage of this method lies in its ability to efficiently establish mappings between brain signals and external stimuli through an optimized alignment mechanism, without the need for manually designed, complex feature extraction steps~\cite{laconte2011decoding}.

Encoder-based alignment models not only address the inefficiency of traditional methods when processing complex brain signals but also enhance the robustness of the model in noisy data environments~\cite{duan2024deep, berahmand2024autoencoders}. Particularly in the context of fMRI signal decoding, such models can significantly improve the reconstruction quality of stimuli from brain signals through precise spatiotemporal feature extraction and cross-modal alignment.

The core idea of encoder-based alignment models is to first extract the spatiotemporal features of fMRI signals using specific encoders, then align these features with external stimuli through a contrastive learning mechanism, achieving effective mapping in a shared feature space. This process consists of two primary steps: spatiotemporal feature extraction and cross-modal alignment.

The key task during the spatiotemporal feature extraction phase is to derive representative spatiotemporal features from the high-dimensional, complex fMRI signals. Given that fMRI signals typically have high dimensionality and contain substantial noise, traditional hand-crafted feature design methods often struggle to capture the key patterns within the signal. Encoder-based models use neural networks (such as convolutional neural networks (CNN), recurrent neural networks (RNN), or Transformer architectures) to perform end-to-end training, automatically extracting effective features from the signals.

\begin{itemize}
    \item \textbf{Convolutional Neural Networks (CNNs)}~\cite{gu2018recent}: CNNs perform local perception on the input signals via convolution kernels, which allows them to automatically learn local spatiotemporal patterns within the signal. In the context of fMRI signal processing, CNNs effectively extract spatial distribution features and capture local brain activity patterns through multiple convolution layers.
    
    \item \textbf{Recurrent Neural Networks (RNNs)}~\cite{schuster1997bidirectional}: Given the sequential nature of fMRI signals, RNNs are particularly adept at handling time-dependent data. RNNs capture temporal dependencies within the signals, thereby learning the dynamic features of brain activity over time.
    
    \item \textbf{Transformer Models}~\cite{vaswani2017attention}: Recently, Transformer architectures have been widely applied to sequential data processing due to their strong global information modeling capabilities. Through self-attention mechanisms, Transformers capture dependencies between different time steps and effectively model the global structure within brain signals.
\end{itemize}

These encoders map the raw fMRI signal \( X_{fMRI} \in \mathbb{R}^{T \times D} \), where \( T \) represents the number of time steps and \( D \) is the feature dimension at each time step, into a latent feature space \( Z_{fMRI} \in \mathbb{R}^{T \times K} \) through a mapping function \( f_{\theta} \), where \( K \) represents the dimension of the latent space and is the output space of the encoder. This mapping process can be expressed as:

\begin{equation}
    Z_{fMRI} = f_{\theta}(X_{fMRI})
\end{equation}

After extracting the spatiotemporal features, the next step involves cross-modal alignment, where the latent representation of the fMRI signal is aligned with the features of external stimuli through contrastive learning, facilitating effective mapping in the shared feature space~\cite{chen2024bridging}.

The goal of cross-modal alignment is to ensure that data from different modalities (such as fMRI signals and stimuli data like images, text, or audio) are mapped to the same latent space, where similar stimuli are close to each other. To achieve this, contrastive learning is commonly used to optimize the model by minimizing the feature distance between the fMRI signal and the external stimulus.

Specifically, for a given latent representation \( Z_{fMRI} \) from the fMRI signal and \( Z_{stimuli} \in \mathbb{R}^{T \times K} \) from the stimulus source, the alignment loss function ensures that they are as close as possible in the latent space. A typical alignment loss function is as follows:

\begin{equation}
    L_{align} = \frac{1}{T} \sum_{t=1}^{T} \| Z_{fMRI,t} - Z_{stimuli,t} \|_2^2
\end{equation}

Here, \( Z_{fMRI,t} \) and \( Z_{stimuli,t} \) represent the fMRI and stimulus features at time step \( t \), respectively, and \( \| \cdot \|_2 \) denotes the L2 norm, which measures the Euclidean distance between the two. The objective of this loss function is to minimize the feature distance between the fMRI signal and the external stimuli, ensuring that they are aligned in the latent space.

The encoder-based model efficiently extracts spatiotemporal features from fMRI signals, particularly when dealing with complex brain signal data. The encoder’s automatic learning mechanism helps identify hidden key patterns, significantly reducing the need for manual feature design. Additionally, during signal mapping, the encoder effectively suppresses noise~\cite{yin2022deep}, enhancing the semantic representation of the signal. This is especially important in low signal-to-noise ratio (SNR) and high-noise fMRI data, as it aids in extracting meaningful signals from complex brain activity. Through a shared latent feature space, the encoder alignment model enables efficient mapping between different modalities~\cite{theodoridis2020cross, lin2020learning, schonfeld2019generalized, arun2022multimodal}, which is crucial for tasks such as cross-modal generation (e.g., fMRI-to-image generation), helping the model learn complex relationships between multimodal data. Compared to large-scale pre-trained generative models, encoder-based alignment models have relatively low computational resource requirements, making them suitable for brain decoding studies in low-resource settings.

Although encoders can extract spatiotemporal features from fMRI signals, effectively fusing features across modalities remains a challenge. The feature differences between different modalities (e.g., fMRI signals are temporal data, whereas images are spatial data) require alignment strategies that can handle these disparities. Furthermore, challenges arise in ensuring semantic consistency and decoding accuracy. Maintaining high semantic consistency between the decoded stimuli and the original stimuli is a critical difficulty for this approach. Designing appropriate loss functions and optimization strategies to ensure that the aligned model can accurately reconstruct the target stimuli is one of the key challenges for encoder alignment models~\cite{yamins2016using}. Despite the encoder's strong performance in high-noise data, the availability of high-quality paired data remains a prerequisite for effective decoding. Data scarcity still limits the widespread application of this method.

To further improve cross-modal alignment, researchers have introduced more complex optimization strategies, such as contrastive loss and triplet loss, in addition to the traditional L2 norm loss. These methods maximize both the similarity and distinguishability between modalities during training, thereby improving alignment performance.

The contrastive loss~\cite{wang2021understanding} function is defined as:

\begin{equation}
    L_{contrastive} = \frac{1}{N} \sum_{i=1}^{N} \left[ \log \frac{ \exp(v_i \cdot t_i / \tau) }{ \sum_{j=1}^{N} \exp(v_i \cdot t_j / \tau) } \right]
\end{equation}

where \( v_i \) and \( t_i \) are the embedding vectors for the image and text, respectively, \( \tau \) is the temperature parameter, and \( N \) is the batch size. Through contrastive learning, the encoder better learns the cross-modal similarities.

A triplet loss function~\cite{cheng2016person} can also be used for optimization:

\begin{equation}
    L_{triplet} = \sum_{i=1}^{N} \left[ \| v_i - t_i \|_2^2 - \| v_i - t_j \|_2^2 + \alpha \right]
\end{equation}

where \( v_i \) and \( t_i \) represent positive sample pairs, \( t_j \) represents a negative sample pair, and \( \alpha \) is a constant parameter, often referred to as the "margin" parameter, which controls the minimum distance between positive and negative sample pairs.

Encoder-based alignment models effectively map the relationship between fMRI signals and external stimuli to a shared latent space through spatiotemporal feature extraction and cross-modal alignment. While these models offer clear advantages in handling high-noise data and multi-modal data alignment, challenges remain in effectively integrating features across modalities, ensuring semantic consistency, and improving the model’s accuracy. Continued optimization of alignment strategies, introduction of new loss functions, and refinement of optimization methods will likely lead to better performance of encoder-based alignment models in brain decoding tasks.

\subsection{Large Language Model-Centric Models}

Large Language Model-Centric Models (LLM-Centric Models) are decoding architectures that leverage large language models (LLMs) as the core component to map fMRI signals into natural language outputs, such as textual descriptions, instruction-based responses, and question-answering inferences~\cite{kuang2024natural}. Unlike end-to-end models, LLM-centric architectures do not rely on direct signal-to-target mapping but instead enhance the semantic representation of fMRI features using the pre-trained knowledge of LLMs~\cite{hu2023survey}, thereby improving decoding quality and interpretability. The key idea behind this approach is to utilize the large-scale semantic space of pre-trained LLMs to achieve multi-level semantic alignment of fMRI signals~\cite{hu2023source}, enabling neural signals to be naturally transformed into contextually coherent textual content. Given the strong reasoning capabilities of LLMs, this architecture not only supports basic semantic descriptions but also facilitates more complex cognitive tasks, such as sentiment analysis and logical reasoning in question-answering. 

In practical applications, LLM-centric architectures typically employ two primary strategies to align fMRI features with the language semantic space: Instruction-Tuning and Multi-Stage Alignment~\cite{chen2024visual}. These methods optimize the decoding capability of fMRI signals through different training strategies, enabling brain activity to be effectively interpreted by LLMs as meaningful linguistic outputs.  

The core idea of the instruction-tuning architecture is to align fMRI features with the semantic space of LLMs through Brain Instruction Tuning (BIT)~\cite{li2025towards}, thereby achieving efficient transformation from brain signals to language. Specifically, this method leverages LLMs for instruction learning, where fMRI features are provided as input during training, along with predefined language tasks (such as text completion and question-answering), allowing LLMs to learn how to generate contextually appropriate textual outputs from fMRI data.  

During the training process, BIT primarily involves the following steps:  
\begin{itemize}
    \item fMRI Encoding: Extracting spatiotemporal features from fMRI signals using convolutional neural networks (CNNs) or Transformer-based models.
    \item Semantic Alignment: Projecting fMRI features into the LLM semantic space to serve as inputs to the LLM.
    \item Language Generation: Generating textual outputs using the LLM and optimizing the loss function to enhance decoding accuracy.
\end{itemize}

The optimization objective of BIT can be expressed as follows:  

\begin{equation}
    L_{BIT} = -\sum \log P_{LLM}(y_t | X_{fMRI}, y_{<t})
\end{equation}

where \( X_{fMRI} \) represents the fMRI input, \( y_t \) denotes the \( t \)-th token generated by the LLM, and \( y_{<t} \) refers to the sequence of tokens generated by the LLM up to time step \( t-1 \). This objective function aims to maximize the alignment between the LLM-generated text sequence and the fMRI signals, thereby improving decoding quality.  

LLM-Centric Models with single-stage alignment support complex reasoning tasks, generate high-quality textual outputs, and facilitate tasks such as question-answering and summarization~\cite{shakil2024abstractive}. Furthermore, these models can fully leverage the pre-trained capabilities of LLMs, reducing dependence on fMRI training data and improving generalization performance. However, LLM-centric models require high-quality brain-language paired datasets, which are costly to annotate and challenging to scale. Additionally, the computational cost of training and inference for LLMs is high, necessitating distributed computing optimization to mitigate resource constraints.  

The core idea of the multi-stage alignment architecture is progressive learning, which first aligns fMRI signals with textual embeddings in a pre-training stage~\cite{liu2025amyloid}, followed by a fine-tuning stage to adapt to specific downstream tasks. The primary advantage of this method is that it leverages large-scale pre-training corpora to establish a more stable mapping between fMRI signals and the LLM semantic space, thereby enhancing model generalization. This approach typically involves the following key steps:  
\begin{itemize}
    \item fMRI Pre-Training: Training an fMRI encoder on large-scale fMRI-text datasets to align its output features with LLM semantic embeddings.
    \item Feature Alignment: Employing contrastive learning, knowledge distillation, or similar techniques to project fMRI features into the LLM semantic space.
    \item Fine-Tuning: Adapting the model to specific downstream tasks to improve decoding performance.
\end{itemize}

The optimization objective of multi-stage alignment typically employs L2 loss for feature alignment:  

\begin{equation}
    L_{align} = \| E_{fMRI}(X) - E_{text}(T) \|^2
\end{equation}

where \( E_{fMRI}(X) \) represents the output features of the fMRI encoder, and \( E_{text}(T) \) denotes the semantic embeddings from the LLM. This optimization ensures that fMRI signal representations closely resemble the text feature representations in the LLM semantic space, thereby improving decoding effectiveness.  

Multi-stage alignment in LLM-centric models enhances adaptability to diverse text generation tasks, improves model generalization, and reduces reliance on a single dataset while increasing robustness to noise in fMRI signals~\cite{shirer2015optimization}. However, this method demands substantial computational resources, making efficient training on standard hardware challenging. Additionally, different tasks may require distinct fine-tuning strategies, increasing implementation complexity.  

Despite the significant potential of LLM-centric architectures in fMRI decoding, several challenges remain. The low temporal resolution and high noise levels of fMRI signals complicate their mapping to linguistic semantics, often leading to information loss during decoding. Moreover, modern LLMs comprise billions or even trillions of parameters, resulting in prohibitively high training and inference costs, necessitating efficient computational optimization techniques. Furthermore, the inherent acquisition delay in fMRI poses challenges for real-time inference, particularly in brain-computer interface (BCI) applications, where optimizing decoding speed remains a critical issue.

\subsection{Hybrid Models}  

Hybrid architectures integrate the advantages of end-to-end models, generative models, and language models to construct a multi-stage, multimodal decoding system capable of high-quality reconstruction of various stimuli, including visual, auditory, and linguistic inputs. Compared to single-method approaches, the core concept of hybrid architectures is to leverage the strengths of different models, optimizing multiple stages—including encoding, generation, and semantic modeling—in a collaborative manner to maximize decoding quality and cross-modal adaptability. Particularly in visual stimulus reconstruction tasks, hybrid architectures harness the powerful modeling capabilities of deep generative models~\cite{ahmed2023deep}, enabling the transformation of fMRI signals into high-quality images and offering a novel paradigm for brain-computer interfaces and neural image decoding.  

Within this framework, the decoding process typically consists of three primary stages: **fMRI feature encoding, modality generation, and semantic modeling**. First, the input fMRI signal undergoes feature extraction via an encoding network to obtain high-level neural representations. Next, a generative model reconstructs the corresponding visual, auditory, or textual information based on these extracted features. Finally, a language model is employed to refine the generated outputs, enhancing their semantic accuracy or directly generating textual descriptions to improve multimodal consistency. A representative implementation of this framework is the encoding-generation joint architecture, which utilizes deep neural networks (DNNs) for fMRI feature extraction~\cite{jang2017task, sze2017efficient}, diffusion models for generating high-quality images, and large language models (LLMs) for textual descriptions, thereby enabling end-to-end cross-modal decoding.  

During the encoding stage, convolutional neural networks (CNNs) or Transformer architectures extract deep features from fMRI signals to capture spatiotemporal patterns in neural activity. Given an fMRI input signal \( X_{fMRI} \in \mathbb{R}^{N \times D} \), where \( N \) denotes the number of time steps and \( D \) represents the feature dimension, CNNs compute convolutional transformations as follows:  

\begin{equation}
    X_{conv} = \sigma(W * X_{fMRI} + b)
\end{equation}

where \( W \) is the convolutional kernel, \( b \) is the bias term, and \( \sigma \) denotes the activation function. The extracted feature vectors are then projected into a latent space \( Z \), which serves as input to the generative model.  

The core of the visual reconstruction stage is the diffusion model, which employs a stepwise denoising strategy to transform random noise into high-quality images. The diffusion process is defined as follows:  

\begin{equation}
    q(\hat{I}_t | \hat{I}_{t-1}) = \mathcal{N}(\hat{I}_t; \sqrt{\alpha_t} \hat{I}_{t-1}, (1 - \alpha_t) I)
\end{equation}

where \( \alpha_t \) controls the noise attenuation rate. The target image \( \hat{I} \) is then recovered using a denoising network trained with the following formulation:  

\begin{equation}
    p_{\theta}(\hat{I}_{t-1} | \hat{I}_t) = \mathcal{N}(\hat{I}_{t-1}; \mu_{\theta}(\hat{I}_t, t), \Sigma_{\theta}(t))
\end{equation}

This diffusion model effectively extracts visual stimulus-related information from fMRI signals, generating high-resolution, semantically consistent images. Following visual reconstruction, the LLM further processes the generated images to provide semantic descriptions, enhancing the interpretability of the decoding process. Given a textual sequence \( Y = \{y_1, y_2, \dots, y_T\} \), the probability of generating text conditioned on fMRI input is computed using a Transformer model as follows:  

\begin{equation}
    P(Y | X_{fMRI}) = \prod_{t=1}^{T} P(y_t | X_{fMRI}, y_{<t})
\end{equation}

This module not only facilitates the generation of visual descriptions but also extends to broader multimodal alignment tasks, such as predicting speech content or musical melodies from fMRI signals, supporting richer perceptual information decoding~\cite{gao2025cinebrain}.  

To ensure coordinated optimization across encoding, generation, and textual parsing components, the loss function of this architecture comprises multiple terms:  

\begin{equation}
    L_{hybrid} = \gamma_1 L_{MSE} + \gamma_2 L_{KL} + \gamma_3 L_{CE}
\end{equation}

where \( L_{MSE} \) optimizes feature learning in the fMRI encoding stage, \( L_{KL} \) constrains the latent distribution of the diffusion model to align more closely with natural image distributions, and \( L_{CE} \) trains the LLM to generate high-quality textual descriptions.  

To address the high noise levels in fMRI signals and the challenge of adapting to cross-site data, hybrid architectures can incorporate noise-robust strategies, such as denoising autoencoders (DAEs)~\cite{majumdar2018blind}, to mitigate interference during data preprocessing and improve signal stability. Given noisy fMRI data \( X_{noisy} \) and its corresponding denoised signal \( X_{clean} \), the DAE optimization objective is defined as follows:  

\begin{equation}
    L_{DAE} = \frac{1}{N} \sum_{i=1}^{N} \| f_{\phi}(X_{noisy,i}) - X_{clean,i} \|^2
\end{equation}

where \( f_{\phi} \) represents the autoencoder’s mapping function. Additionally, this approach incorporates a multimodal fusion module to integrate other neural signals, such as EEG and MEG, thereby enhancing the system’s ability to decode perceptual information. The fusion module is optimized using the following objective:  

\begin{equation}
    L_{MM-Fusion} = \| E_{fMRI}(X_{fMRI}) - E_{EEG}(X_{EEG}) \|^2
\end{equation}

where \( E_{fMRI} \) and \( E_{EEG} \) denote the encoding networks for their respective modalities.  

While hybrid architectures exhibit significant advantages in enhancing multimodal reconstruction capabilities—particularly in high-quality visual stimulus reconstruction—combining diffusion models with Transformer-based architectures enables the generation of highly detailed and structurally coherent images. However, several challenges remain. First, the multi-stage processing pipeline, which involves CNNs, diffusion models, and LLMs, results in high computational complexity and substantial training costs, making real-time applications difficult. Second, strict feature alignment between different modalities (e.g., fMRI, EEG, images, and text) is essential; otherwise, reconstruction quality may be compromised. Additionally, given the limitations of dataset size and quality, further improvements in the generalization ability of hybrid architectures are necessary to enable broader applicability. Future research directions may focus on efficient optimization techniques, self-supervised learning strategies, and lightweight model designs to enhance the scalability of hybrid architectures, allowing them to play a more significant role in neural image decoding, brain-computer interfaces, and cognitive science.

\section{Evaluation Metrics}\label{sec:evaluation metrics}

To evaluate the similarity between the ground truth stimuli and their reconstructions, two key aspects are examined: detail fidelity and semantic fidelity. Therefore, the metrics can be grouped into high-level and low-level.

\subsection{Quick Overview}
To better understand the definition of different metrics, we provide a quick overview in this section. 

\noindent{\textbullet~Low-Level Evaluation Metrics} measure image features that capture \textit{essential information} about the visual content and structure of the image. Examples include PixCorr, SSIM, AlexNet(2), and AlexNet(5). 

\noindent{\textbullet~High-Level Evaluation Metrics} assess images based on their \textit{semantic information}, such as object relationships, and contextual understanding. Examples include Incep, CLIP, EffNet-B and SwAV. 

\noindent{\textbullet~Evaluation Metrics for Other Modalities} given that research on decoding speech and video stimuli is relatively limited, there is no widely accepted standard for evaluation metrics. Therefore, the description of evaluation metrics for these two modalities is provided in Sec.~\ref{sec:quantitative results}.

\subsection{Definition Details}\label{sec:evaluation metrics sec}

\subsubsection{Low-Level Evaluation Metrics}

\noindent{\textbullet~\textbf{PixCorr}\cite{quan2024psychometry}} measures the pixel-wise correlation between the ground truth $I\in\mathbb{R}^{W\times H}$ and reconstructed image $R\in\mathbb{R}^{W\times H}$. Each image is first flattened into a one-dimensional vector.
\begin{equation}
\begin{aligned}
I^{\prime} = \text{Flatten}(I), 
R^{\prime} = \text{Flatten}(R),
\end{aligned}
\label{Pixelwise}
\end{equation}
where $I^{\prime}$ and $R_i^{\prime}$ are the corresponding flattened vectors. 
The PixCorr score is the correlation coefficient:
\begin{equation}
\text{PixCorr}  = \frac{\text{Cov}(I^{\prime}, R^{\prime})}{\sigma(I_i') \sigma(R^{\prime})},
\label{PixCorr}
\end{equation}
where $\text{Cov}(I^{\prime}, R^{\prime})$ is the covariance between $I^{\prime}$ and $R^{\prime}$, and $\sigma(I^{\prime})$ and $\sigma(R^{\prime})$ are their standard deviations. 

\noindent{\textbullet~\textbf{SSIM}~\cite{wang2004image}} is the Structural Similarity Index Measure, which quantifies image similarity by comparing brightness, contrast, and structural information. For each image pair, SSIM is defined as:
\begin{equation}
\label{SSIM1}
\text{SSIM} = \frac{(2\mu_I \mu_R + C_1)(2\sigma_{IR} + C_2)}{(\mu_I^2 + \mu_R^2 + C_1)(\sigma_I^2 + \sigma_R^2 + C_2)}
\end{equation}
where $ \mu_I $ and $ \mu_R $ are the local means, $ \sigma_I $ and $ \sigma_R $ are the local variances of images $ I $ and $ R $, and $ \sigma_{IR} $ is their covariance. $ C_1 $ and $ C_2 $ are constants to stabilize the calculation.

\noindent{\textbullet~\textbf{AlexNet}~\cite{krizhevsky2012imagenet}} evaluate image reconstruction quality in feature space on AlexNet. It first extracts feature representations at different layers for both ground truth and reconstructed images. 
Features are extracted from the 2nd and 5th layers of AlexNet for each image pair:
\begin{equation}
\label{AlexNet1}
F_{(I,i)} = \text{AlexNet}(I_i, \text{layer}), \quad F_{(R,i)} = \text{AlexNet}(R_i, \text{layer})
\end{equation}
where $ F_{(I,i)} $ and $ F_{(R,i)} $ are feature representations of the ground truth and reconstructed images at the specified layer. 
For each image pair, the similarity of the extracted features is calculated. The percent of correctly matched pairs based on a similarity threshold is defined as:
\begin{equation}
\label{AlexNet2}
\text{Percent Correct} = \frac{1}{N} \sum_{i=1}^N \text{Indicator}_i,
\end{equation}
where $\text{Indicator}_i$ is a binary indicator that marks whether the image pair $i$ meets the similarity threshold. 
The average accuracy for each layer is then computed:
\begin{equation}
\label{AlexNet3}
\text{AlexNet}_\text{layer} = \frac{1}{N} \sum_{i=1}^N \text{Percent Correct}_{(i, \text{layer})}
\end{equation}

\subsubsection{High-Level Evaluation Metrics}
\noindent{\textbullet~\textbf{EffNet-B~\cite{tan2019efficientnet,szegedy2017inception}}} evaluate the latent distance, which measures the similarity between the artificial neural network's response and the brain's mechanisms for core object recognition. 
Specifically, EfficientNet-B1 and ResNet50~\cite{he2016deep} are used to extract image features from the ground truth and reconstructed images, respectively. The average correlation distance between these features is then calculated. Pearson Correlation: Measures the correlation between feature vectors $ X $ and $ Y $ for ground truth and reconstructed images.
\begin{equation}
r = \frac{\text{Cov}(X, Y)}{\sigma_X \sigma_Y}
\label{eq:PearsonCorrelation}
\end{equation}
where $ \text{Cov}(X, Y) $ is the covariance between $ X $ and $ Y $, and $ \sigma_X $ and $ \sigma_Y $ are their standard deviations. Average Correlation Distance: This is the mean correlation distance across all sample pairs.
\begin{equation}
\label{Average Correlation Distance}
\text{Average Correlation Distance} = \frac{1}{N} \sum_{i=1}^N (1 - r)
\end{equation}
where $ N $ is the number of sample pairs, and $\text{Pearson Correlation}$ is the Pearson correlation coefficient for each pair.

\noindent{\textbullet~\textbf{\textbf{SwAV}~\cite{caron2020unsupervised}}} measures the SwAV-ResNet-50 distance between generated images and the ground truth. 

\noindent{\textbullet~\textbf{\textbf{CLIP}~\cite{radford2021learning}}} CLIP is used to extract textual features from ground truth and predicted (reconstructed) stimuli using a pre-trained CLIP model and tokenizer. The cosine similarity between these textual features is then calculated to yield the CLIP similarity score.
\begin{equation}
\label{Cosine Similarity}
\text{Cosine Similarity}(A, B) = \frac{A \cdot B}{\|A\| \|B\|}
\end{equation}
where $A$ and $B$ are textual features extracted for the ground truth and prediction.

\noindent{\textbullet~\textbf{Incep}~\cite{szegedy2016rethinking}} considers the KL divergence between the generated image and ground truth in feature space extracted by the Inception V3 model~\cite{szegedy2016rethinking}:
\begin{equation}
\text{Incep} = \exp \mathbb{E}_{x \sim p} \, KL\left( p(y|x) \parallel p(y) \right),
\end{equation}
where $p(y)$ denotes the average class distribution of the images generated by $G$ in the output layer of the Inception model. A higher Incep score indicates better quality of the generated image.

\section{Experiments}\label{sec:experiments}

The primary objective of fMRI-based brain decoding tasks is to reconstruct external stimuli from fMRI signals passively generated in response to these stimuli. The outcomes of this process require careful attention to the similarity between the ground truth and the reconstructed information. In this section, we will introduce the main evaluation metrics and present both qualitative and quantitative results of some baseline models that we have replicated. All experimental results will be provided under the same dataset and hardware conditions to serve as a reference for future research.

\input{tabels/results}

\input{tabels/results_video}

\subsection{Qualitative Results}\label{sec:qualitative results}

\begin{figure*}[t!]
\centering
\includegraphics[width=0.75\linewidth]{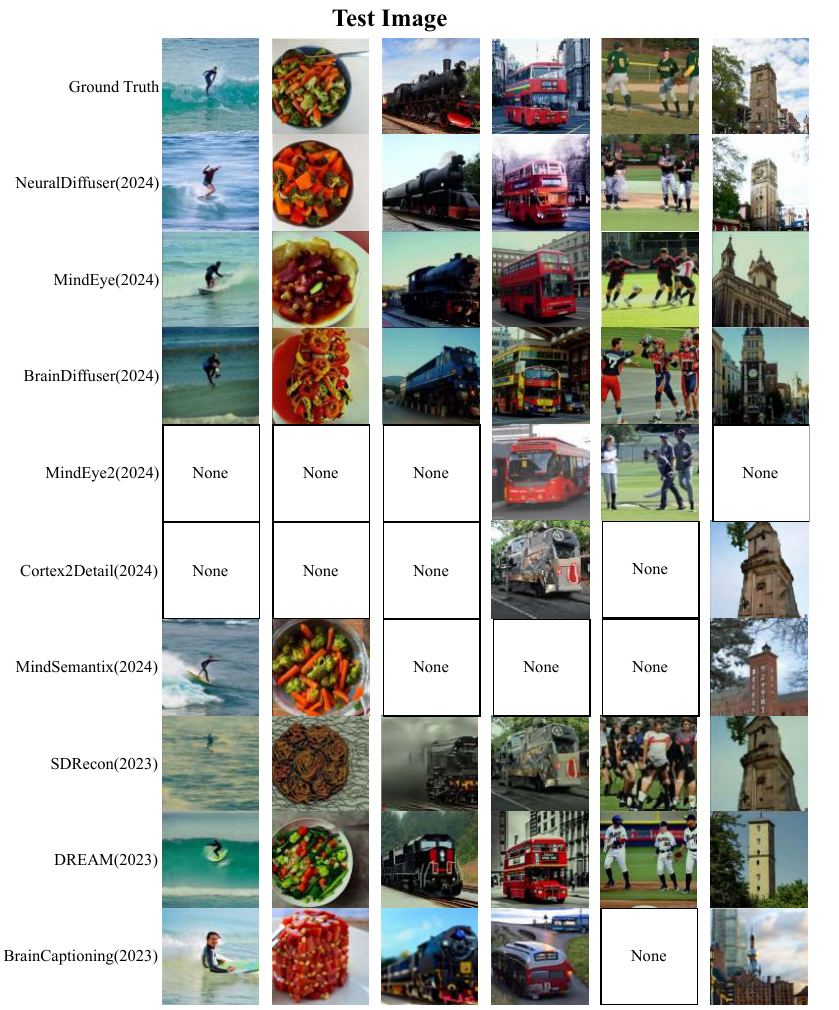}
\caption{The basic structure of fMRI-based brain decoding tasks: After the human body receives external stimuli, the brain generates signals in specific regions. By decoding these brain signals and using generative models for reconstruction, the stimulus signals received by the brain are restored.}
\label{fig:Image_NSD}
\end{figure*}

For image reconstruction decoding, conventional algorithms typically generate images using a fixed seed. Qualitative results for all fMRI-based brain decoding image reconstruction tasks are shown in the figure~\ref{fig:Image_NSD}. The reconstructed content was selected from representative studies and presented alongside the corresponding ground truth. The results primarily focus on the NSD dataset. Specifically, the first row of Figure~\ref{fig:Image_NSD} displays a subset of image stimuli from the NSD dataset. Each subsequent row represents reconstruction results of these stimuli by models selected from corresponding papers or open-source codes. Some reconstruction results were not displayed in the original papers, and therefore, they are not included in the figure~\ref{fig:Image_NSD}. 

For all I2I tasks, the models demonstrated high semantic fidelity (e.g., surfing, food, trains) as well as preservation of details such as color, main object features, and contours. However, the generated images still exhibit certain semantic and detail-related errors and omissions. For instance, MindEye generates high-quality images but suffers from semantic mismatches (e.g., incorrect object categories) and detail inconsistencies (e.g., illogical object appearances). Similarly, images reconstructed by SDRecon are relatively blurry and also face issues with semantic mismatches.

\subsection{Quantitative Results}\label{sec:quantitative results}

To facilitate a quantitative comparison of the current models, we summarize the evaluation metrics for both low-level and high-level aspects used in previous studies in Section~\ref{sec:evaluation metrics sec}. Table~\ref{tab:results} highlights several representative works and their corresponding results on the NSD, GOD, and BLOD5000 datasets. These results are extracted from the respective papers and open-source codes. Due to differences in experimental setups, we do not provide direct performance comparisons between models, as this would be unfair. 

For instance, even within the same dataset, BrainCaptioning uses data from subjects 1 and 2, while BrainDiffuser selects subjects 1, 3, and 5, and MindEye2 uses data from subjects 1 to 8. For most models, the data from subjects 1, 2, 5, and 7 are typically used. Given such experimental settings, we have not provided a dedicated column in Table~\ref{tab:results}'s "Note" section to list these details explicitly.

In addition, the evaluation of video and image reconstruction mainly differs in the temporal dimension. Video information includes both temporal and spatial data, and capturing the temporal changes in the reconstructed video is primarily achieved through motion vectors and optical flow. In Table~\ref{tab:resultsvideos}, we present the low-level and high-level evaluation metrics for video reconstruction on the DNV video reconstruction dataset and provide a comparative analysis of the results. These results are extracted from the corresponding original papers.

Similar to image reconstruction evaluation metrics, SSIM, MSE, and PSNR (Peak Signal-to-Noise Ratio)~\cite{korhonen2012peak} are used as low-level metrics for video reconstruction, while ACC is employed as a high-level metric. The Mean Squared Error (MSE) is defined as:

\begin{equation}
    \text{MSE} = \frac{1}{N} \sum_{i=1}^{N} (I_{\text{reconstructed},i} - I_{\text{groundtruth},i})^2
\end{equation}

where \( I_{\text{reconstructed},i} \) and \( I_{\text{groundtruth},i} \) represent the pixel values of the reconstructed and ground truth video frames, respectively, and \( N \) is the total number of pixels. A lower MSE indicates a closer match between the reconstructed video and the ground truth.

The PSNR, a common metric for evaluating image and video quality, is defined as:

\begin{equation}
    \text{PSNR} = 10 \log_{10}\left( \frac{I_{\text{max}}^2}{\text{MSE}} \right)
\end{equation}

where \( I_{\text{max}} \) represents the maximum possible pixel value (e.g., 255 for an 8-bit image). Higher PSNR values typically correspond to better quality reconstructions, indicating lower levels of distortion.

Additionally, Rank scores are determined by calculating the similarity between reconstructed video segments and ground truth video segments, with the ranking based on \( n \) segments (one correct segment and \( n-1 \) distractor segments). The computed similarity is ranked, and scores are assigned based on the position of the correct ground truth segment in the list. A lower score indicates higher similarity between the two video segments.

These evaluation metrics primarily assess the spatial consistency and object consistency between the reconstructed video and ground truth video. However, given the complex temporal and spatial information in videos, additional metrics such as Multi-Scale SSIM (MS-SSIM)~\cite{wang2003multiscale} and Spatio-Temporal SSIM (ST-SSIM)~\cite{moorthy2010efficient} can be introduced to better measure the reconstruction quality. Furthermore, for specific categories of reconstructed videos, the videos can be applied to downstream tasks (e.g., action recognition~\cite{guo2024benchmarking,li2024prototypical}, action detection~\cite{liu2024micro}) and evaluated using task-specific performance metrics (e.g., accuracy, recall) to assess the quality of the reconstructed video.

For evaluating audio reconstruction, several metrics are employed to assess both low-level and high-level characteristics of the reconstructed sound. Specifically, BSR (Binaural Signal-to-Noise Ratio), ACC (Accuracy), PCC (Pearson Correlation Coefficient), and Mel-Cepstral Distortion (MCD) are used as key evaluation metrics. Furthermore, due to the limited number of existing studies on speech reconstruction and the lack of a unified evaluation framework, we do not provide a comparative analysis of related works here.

The Binaural Signal-to-Noise Ratio (BSR)~\cite{hawkins1984signal} measures the perceptual quality of the audio signal and is defined as:

\begin{equation}
    \text{BSR} = 10 \log_{10} \left( \frac{\sum_{i=1}^{N} |S_{\text{re}, i}|^2}{\sum_{i=1}^{N} |S_{\text{noise}, i}|^2} \right)
\end{equation}

where \( S_{\text{re}, i} \) and \( S_{\text{noise}, i} \) represent the reconstructed audio signal and noise components in the frequency domain, respectively, and \( N \) is the number of audio samples. A higher BSR value indicates that the reconstructed audio signal has a higher signal-to-noise ratio, suggesting better quality reconstruction.

Accuracy (ACC) is another important metric, especially in tasks such as speech recognition or classification of audio events. It is defined as the proportion of correctly predicted audio segments compared to the total number of segments:

\begin{equation}
    \text{ACC} = \frac{\text{Correctly Predicted Segments}}{\text{Total Segments}}
\end{equation}

The Pearson Correlation Coefficient (PCC)~\cite{cohen2009pearson} quantifies the linear relationship between the reconstructed audio and the ground truth. It is computed as:

\begin{equation}
    \text{PCC} = \frac{\sum_{i=1}^{N} (S_{\text{re}, i} - \overline{S_{\text{re}}})(S_{\text{gt}, i} - \overline{S_{\text{gt}}})}{\sqrt{\sum_{i=1}^{N} (S_{\text{re}, i} - \overline{S_{\text{re}}})^2 \sum_{i=1}^{N} (S_{\text{gt}, i} - \overline{S_{\text{gt}}})^2}}
\end{equation}

where \( \overline{S_{\text{re}}} \) and \( \overline{S_{\text{gt}}} \) are the mean values of the reconstructed and ground truth audio signals, respectively. A higher PCC indicates a stronger correlation between the reconstructed and the ground truth audio signals, which generally corresponds to better reconstruction quality.

Mel-Cepstral Distortion (MCD) is a metric used to measure the distortion between two audio signals in the Mel frequency cepstral domain. The MCD is computed as:

\begin{equation}
    \text{MCD} = \frac{1}{N} \sum_{i=1}^{N} \left( \sum_{j=1}^{M} \left( C_{\text{re}, ij} - C_{\text{gt}, ij} \right)^2 \right)^{1/2}
\end{equation}

where \( C_{\text{re}, ij} \) and \( C_{\text{gt}, ij} \) are the Mel-cepstral coefficients of the reconstructed and ground truth audio signals, respectively, and \( N \) is the total number of frames, while \( M \) represents the number of coefficients per frame. Lower MCD values indicate that the reconstructed audio has less distortion from the original signal, leading to higher-quality audio reconstruction.

These metrics—BSR, ACC, PCC, and MCD—offer a comprehensive assessment of the performance of audio reconstruction models, accounting for both perceptual quality and the accuracy of the reconstruction. Each metric provides a unique perspective, from signal quality and correlation to perceptual distortion, making them essential for evaluating models that aim to reconstruct complex audio signals such as speech or environmental sounds.

\section{Future Directions}\label{sec:future}

In recent years, significant advancements have been made in fMRI-based brain decoding. While these methods have achieved notable results to some extent, there remains a need for substantial progress in areas such as efficient decoding models, noise handling, individual differences, multimodal integration, real-time or continuous decoding, personalized interaction, paper-related research, and distributed computing. We have identified several promising research directions:

\subsection{High-Resolution Temporal Dynamics Modeling}\label{sec:FD1}  
The low temporal resolution of current fMRI data poses significant challenges in capturing rapid changes in brain activity, especially in cognitive processes such as vision and hearing. In addition to combining fMRI with hybrid neuroimaging methods, such as EEG and MEG, which offer higher temporal resolution, efforts can also be made to improve the accuracy of temporal dynamics modeling. Current neural network models mostly focus on spatial feature extraction, with limited attention to temporal modeling. Therefore, models with temporal sequence modeling capabilities, such as Long Short-Term Memory Networks (LSTM)~\cite{cheng2016long} or 3D Convolutional Neural Networks (3D CNN)~\cite{ji20123d}, can be enhanced to better integrate temporal and spatial information. These models would enable more effective decoding of rapidly changing brain features, leading to more detailed temporal dynamics reconstruction. Future research could also draw on super-resolution reconstruction~\cite{zhang2022cross, chen2024high} techniques from the image domain, using algorithms to compensate for the temporal blurring in fMRI data~\cite{eklund2012fmri}. A dedicated network could be designed to learn the mapping between low-resolution and high-resolution temporal sequences~\cite{schultz1996extraction, isobe2020video, cao2021video}, thereby enriching low-resolution fMRI signals to improve the model’s ability to capture rapidly changing neural activity. Additionally, beyond integrating various neuroimaging signals, exploring synchronous analysis of other biological signals with fMRI could further enhance brain activity capture across different timescales via multimodal data fusion.

\subsection{Personalized Brain Decoding Frameworks}\label{sec:FD2}  
Individual differences in cognition and decoding remain a major challenge in future brain decoding research, particularly in visual perception, where individuals may perceive stimuli differently. One future direction will be to adapt decoding models based on individual characteristics. In addition to anatomical differences, an individual’s psychological state (e.g., emotional fluctuations, cognitive states) and physiological features (e.g., heart rate) also influence brain activity~\cite{koban2021self, candia2022cardiac}. Therefore, personalized decoding models in the future will need to integrate multimodal data, including physiological and behavioral data, through fusion learning techniques, to achieve more precise and accurate personalized brain decoding. For individual differences, techniques such as Transfer Learning~\cite{zhu2023transfer} and Meta-Learning~\cite{vettoruzzo2024advances} hold significant potential. Transfer learning can apply general features learned from large-scale population data to individual cases, while Meta-Learning enables rapid adaptation to new individuals with minimal data. Combining these methods could lead to more personalized brain decoding models.

\subsection{Multimodal Neural-Semantic Alignment}\label{sec:FD3}  
When the brain processes visual information, there are two types of signal interactions: bottom-up perceptual signals (e.g., raw visual data from sensory organs like the eyes) and top-down semantic predictions (e.g., inferences based on prior experiences, knowledge, and context)~\cite{hardstone2021long}. These two signals interact to help us quickly and efficiently recognize objects, scenes, or contexts. Future research could enhance decoding capability by further aligning multimodal neural and semantic information. Current research generally processes visual and linguistic data separately. Future work could focus on joint embedding spaces, tightly coupling fMRI-based neural activity representations with semantic information from textual descriptions or image labels~\cite{he2024fm}. In this way, the model could not only comprehend visual information but also align it with semantic data to more accurately reconstruct the stimulus content. Moreover, visual-semantic alignment should not be limited to a single layer of information. For instance, low-level features of a visual stimulus (e.g., color, edges) and high-level features (e.g., category, emotional significance) may show distinct processing patterns and response strengths in the brain. Therefore, aligning and mapping these low-level and high-level features could provide a more comprehensive integration of visual-semantic information, fully utilizing the data. Since visual perception is often accompanied by emotional responses, future multimodal decoding models could treat emotion labels as an additional semantic dimension, incorporating these labels into the neural network to enrich the decoding of brain signals~\cite{liu2023emotion}.

\subsection{Real-Time Adaptive Decoding}\label{sec:FD4}  
With the rise of brain-computer interfaces (BCIs)~\cite{tang2023flexible}, real-time brain decoding has become increasingly important. To achieve real-time, precise decoding, there is a need for computationally efficient models with low memory usage. Techniques such as model quantization can reduce computational demands, enabling rapid real-time inference. Additionally, the network structure and performance can be dynamically adjusted based on device capabilities to ensure good performance on different hardware platforms. Real-time decoding requires networks capable of handling the non-stationary nature of neural signals. Current datasets are often collected in controlled laboratory environments, where brain signals exhibit minimal fluctuation over short periods. However, in real-world conditions, brain signals may fluctuate dramatically due to both internal and external influences. To address this, the decoding system could adopt online learning mechanisms, enabling the model to continuously update during the decoding process, thereby adapting to changes in an individual's neural activity. This could be achieved through incremental learning or reinforcement learning, allowing the model to optimize its decoding capacity in real time. In BCI applications, real-time feedback is crucial, and further exploration of closed-loop systems—where stimulus presentation is dynamically adjusted based on decoded information—could enhance cognitive or behavioral states. For example, in neurofeedback training~\cite{khan2025real, pan2022review}, real-time decoding of brain activity via BCIs could adjust visual stimuli to guide brain activity toward a pre-defined target state, facilitating self-regulation in a closed-loop manner.

\subsection{Ethical and Security Considerations}\label{sec:FD5}  
As brain decoding technology becomes more widely applied, ethical concerns have emerged as a critical issue~\cite{davis2017brain}. Personal brain data may leak sensitive information, requiring stricter data encryption and privacy protection mechanisms to ensure data security and privacy~\cite{shamoo2010ethical, fenton2009ethical, shen2024ethical, mecacci2019identifying}. Furthermore, data from different populations or cognitive backgrounds may introduce biases in decoding models. If a model is trained on data from a specific group, it may not generalize well to others~\cite{liu2021just, tian2025brainguard}. Researchers must focus on eliminating biases in models by designing and collecting more diverse, fair datasets, ensuring that the decoding model performs consistently across individuals~\cite{parraga2023fairness}. Additionally, since brain decoding technology may be widely used in clinical settings, errors in the model’s results could have serious implications for clinical diagnoses. Therefore, interpretable decoding models are crucial. Interpretable results help doctors and patients understand the process, enhancing the model’s credibility and usability in clinical applications.

\subsection{Clinical Translation and Real-World Validation}\label{sec:FD6}  
While brain decoding models have demonstrated groundbreaking progress in laboratory settings, applying these models in clinical and real-world conditions still poses significant challenges. fMRI scanners may vary across different medical institutions, laboratories, and real-world environments, complicating cross-institutional and cross-patient group decoding. Therefore, future research should focus on standardized data processing and normalization techniques to ensure seamless integration of data from different devices, improving the model’s adaptability across various settings. Clinical brain decoding goes beyond diagnostics; it can also aid medical treatments~\cite{yin2022deep}. For example, real-time decoding of visual information can help stroke patients by providing visual feedback for rehabilitation training. Furthermore, future research could explore the application of brain decoding in neuro-rehabilitation and psychotherapy, providing more personalized treatment options for patients. Brain decoding combined with neurofeedback training can help patients self-regulate their brain activity~\cite{krause2021self} to improve symptoms of conditions like ADHD or anxiety. In such cases, the decoding model can detect the patient’s brain activity in real time and provide feedback to facilitate self-regulation.

\section{Conclusion}\label{sec:conclusion}

This survey provides a systematic review and summary of both the foundational research and recent advances in the field of fMRI-based brain decoding. We primarily focus on reconstruction tasks utilizing functional magnetic resonance signals, offering a comprehensive introduction to relevant datasets and key brain regions involved in these processes. 
In addition, we classify and organize existing models, outlining their primary motivations and key features. Subsequently, we present both qualitative and quantitative analyses of several well-known models under a unified hardware environment, sharing our findings to establish baseline models for future research. 
This survey serves as an invaluable resource for researchers in this field, offering valuable insights that may guide further development and progress in brain decoding research.

\section*{Acknowledgment}

This work was supported by the National Key Research and Development Program of China under Grant No.2022YFF1202400.(Corresponding author: Guohua Dong and Xiaomin
Ying.)

\ifCLASSOPTIONcaptionsoff
\newpage
\fi

\clearpage
{\small
\bibliographystyle{IEEEtran}

\bibliography{IEEEabrv, main}
}

\end{document}

%% file: tabels/reviewsandsurveys.tex
\begin{table}[]
\caption{Reviews and Surveys}
\renewcommand\arraystretch{1.4}
\resizebox{1.0\linewidth}{!}{
\begin{tabular}{|l|l|l|l|}
\hline
\thickhline
Title & Year & Venue & Description \\ \hline
\makecell[l]{
Du~\etal~\cite{du2019brain}} & 2019 & Engineering & \makecell[l]{Reviewed solutions prior to 2019, emphasizing early unified \\ encoder-decoder frameworks and the positive impact of \\ deep generative models on brain encoding and decoding.}\\ \hline

\makecell[l]{
Cao~\etal~\cite{cao2021computational}} & 2021 & IJCAI & \makecell[l]{Focused on studies that use human language as the primary \\ stimulus source, exploring the unique challenges and \\ methodologies of this domain.}\\ \hline

\makecell[l]{
Oota~\etal~\cite{oota2023deep}} & 2023 & arXiv & \makecell[l]{Specifically investigated fMRI-based brain encoding and \\ decoding architectures, providing an overview of \\ relevant datasets and fundamental model structures.}\\ \hline

\makecell[l]{
Mai~\etal~\cite{mai2023brain}} & 2023 & arXiv & \makecell[l]{Surveyed brain decoding models up to 2023, covering \\ methods that reconstruct stimuli from \\ various types of brain signals.}\\ \hline

\makecell[l]{
Zhao~\etal~\cite{zhao2025diffusionmodelscomputationalneuroimaging}} & 2025 & arXiv & \makecell[l]{Computational neuroimaging supports cognition and behavior \\ research by analyzing brain images or signals, \\ and this survey reviews the advancements in the application of \\ diffusion models to neuroimaging tasks.}\\ \hline

\end{tabular}}
\label{tab:reviewsandsurveys}
\end{table}

%% file: tabels/datasets.tex
\begin{table*}[t]
\centering
\caption{Datasets for decoding brain activity into various forms. I2I denotes Image-to-Image, I2T represents Image-to-Text, S2S refers to Speech-to-Speech, S2T signifies Speech-to-Text, V2V stands for Video-to-Video, and V2T denotes Video-to-Text.}
\renewcommand\arraystretch{2}
\resizebox{1.0\textwidth}{!}{
\begin{tabular}{|c<{\centering}|c<{\centering}|c<{\centering}|c<{\centering}|c<{\centering}|c<{\centering}|c<{\centering}|c<{\centering}|c<{\centering}|}
\hline
\thickhline


\multirow{2}{*}{\textbf{Modality}} & \multirow{2}{*}{\textbf{Dataset}} & \multirow{2}{*}{\textbf{Type}} & \multirow{2}{*}{\textbf{Year} \textbf{\&} \textbf{Publish}} & \multicolumn{2}{c|}{\textbf{Experimental Paradigm}} & 
\multirow{2}{*}{\textbf{Subject}} & \multirow{2}{*}{\textbf{ROI}} & \multirow{2}{*}{\textbf{Brain} Task} \\ \cline{5-6} & & & & 
\multicolumn{1}{c|}{\textbf{Stimuli}} & \textbf{Stimuli Source} & & & \\ \hline

\multirow{13}{*}{Image}  

& Vim-1~\cite{kay2008identifying} & fMRI & 2008, Nature & \multicolumn{1}{c|}{1870(1750+120)}     
& {\makecell{Commercial Digital Library \&\\ Berkeley Segmentation Dataset}} & 2 & V1, V2, V3, V4 & I2I \\ \cline{2-9} 

& BCP~\cite{miyawaki2008visual} & fMRI & {\makecell{2008, \\ Neuron}} & \multicolumn{1}{c|}{}         & Composite Image & 4 & V1, V2, V3 & I2I \\ \cline{2-9}

& HWD~\cite{van2010neural} & fMRI & {\makecell{2010, \\ Neural computation}} & \multicolumn{1}{c|}{2106}         & MNIST & 1 &  & I2I \\ \cline{2-9}

& Brains~\cite{schoenmakers2013linear} & fMRI & {\makecell{2013, \\ NeuroImage}} & \multicolumn{1}{c|}{720}         & Handwritten Characters & 3 & V1 & I2I \\ \cline{2-9}

& HCP~\cite{van2013wu} & fMRI & 2013, Neuroimage & \multicolumn{1}{c|}{} &  
& 1200 & & \\ \cline{2-9} 

& GOD~\cite{horikawa2017generic} & fMRI & {\makecell{2017,\\Nature communications}}& \multicolumn{1}{c|}{1250}
& ImageNet & 5 & {\makecell{V1, V2, V3, V4,\\LOC, FFA, PPA}}  & I2I, I2T \\ \cline{2-9} 

& MSC~\cite{gordon2017precision} & fMRI & 2017, Neuron & \multicolumn{1}{c|}{} & 
& 10 & & \\ \cline{2-9} 

& EEG-VOA~\cite{spampinato2017deep} & EGG & 2017, CVPR & \multicolumn{1}{c|}{2000} & ImageNet
& 6 & 128 EEG Channels & I2I \\ \cline{2-9} 

& DIR~\cite{shen2019deep} & fMRI & {\makecell{2019,\\PLoS computational biology}} & \multicolumn{1}{c|}{50} 
& {\makecell{Artificial shapes \&\\alphabetical letters}} & 3 & {\makecell{V1, V2, V3, V4,\\LOC, FFA, PPA}} & I2I \\ \cline{2-9} 

& CelebrityFace~\cite{vanrullen2019reconstructing} & fMRI & {\makecell{2019,\\Communications biology}} & \multicolumn{1}{c|}{108(88+20)} & CelebA                       & 4 & TL, OL, PL, FL & I2I \\ \cline{2-9} 

& BOLD5000~\cite{chang2019bold5000} & fMRI & {\makecell{2019, \\ Scientific data}} & \multicolumn{1}{c|}{4916}         & SUN, COCO, and ImageNet & 4 & PPA, RSC, OPA, EVC & I2I, I2T \\ \cline{2-9} 

& OCD~\cite{huang2020long} & fMRI & {\makecell{2020,\\Human brain mapping}} & \multicolumn{1}{c|}{{\makecell{2750\\(2250+250+250)}}} & ImageNet 
& 5 & {\makecell{V1, V2, V3, LVC,\\HVC, VC}} & I2T \\ \cline{2-9} 

& NSD~\cite{allen2021massive} & fMRI & 2021, ArXiv & \multicolumn{1}{c|}{{\makecell{70566\\(8×9000+1000)}}} 
& COCO& 8 & {\makecell{V1, V2, V3, V4,\\LOC, FFA, PPA}} & I2I, I2T \\ 

\hline

\multirow{8}{*}{Sound}    

& MusicAffect~\cite{daly2020neural} & fMRI & 2020, Scientific data & \multicolumn{1}{c|}{} & {\makecell{Synthetic music \&\\ Classical music clips}}                      
& 21 & & S2S \\ \cline{2-9} 

& Story~\cite{nastase2020leveraging} & fMRI & 2020, NeuroImage & \multicolumn{1}{c|}{10} & Story                                                                  
& 149 & {\makecell{EAC, AAC, \\TPOJ, PMC}} & S2S \\ \cline{2-9} 

& Narratives~\cite{nastase2021narratives} & fMRI & 2021, Scientific data & \multicolumn{1}{c|}{27} & Story                                                                
& 345 & {\makecell{A1, Mbelt, LBelt,\\Pbelt, RI}} & S2T \\ \cline{2-9} 

& MusicGenre~\cite{nakai2022music} & fMRI & 2022, Data in Brief & \multicolumn{1}{c|}{540(480+60)} & Music                                                                
& 3 & & S2S \\ \cline{2-9} 

& BSR~\cite{park2023sound} & fMRI & 2023, ArXiv & \multicolumn{1}{c|}{1250(1200+50)} & VGGSound test dataset                                                  
& 5 & {\makecell{A1, LBelt, Pbelt,\\A4, A5}} & S2S \\ \cline{2-9} 

& ETCAS~\cite{guo2023end} & EGG & {\makecell{2023,\\Knowledge-Based Systems}} & \multicolumn{1}{c|}{} & TIMIT dataset                                                     
& 50 & 24 EEG Channels & S2S \\ \cline{2-9} 

& GTZan~\cite{ferrante2024r} & fMRI & 2024, ArXiv & \multicolumn{1}{c|}{540} & Music                                                                  
& 5 & & S2S \\

\hline

\multirow{9}{*}{Video}    
& VER~\cite{nishimoto2011reconstructing} & fMRI & 2011, Current biology & \multicolumn{1}{c|}{12600} & {\makecell{Apple Quick-Time gallery \&\\ Youtube}}                 
& 3 & {\makecell{V1, V2, V3,\\V3A, V3B}} & V2V \\ \cline{2-9} 

& Cam-CAN~\cite{taylor2017cambridge} & fMRI & 2017, Neuroimage & \multicolumn{1}{c|}{8m40s+8m40s} & Movie clips                                                           
& 656 & & V2V \\ \cline{2-9} 

& DNV~\cite{wen2018neural} & fMRI & 2018, Cerebral cortex & \multicolumn{1}{c|}{972(374+598)} & Video Blocks \& Youtube                                                
& 3 & {\makecell{V1, V2, V3, V4,\\LO, MT,FFA, PPA,\\ LIP, TPJ, PEF\\}} & V2V \\ \cline{2-9} 

& STNS~\cite{seeliger2019large} & fMRI & 2019, BioRxiv & \multicolumn{1}{c|}{30} & Doctor Who                                                        
& 1 & {\makecell{V1, V2, V3, MT,\\AC, FFA, LOC, OFA}} & V2V \\ \cline{2-9} 

& NNDB~\cite{aliko2020naturalistic} & fMRI & 2020, Scientific Data & \multicolumn{1}{c|}{10} & Full movies                                                
& 86 & & V2V \\ \cline{2-9} 

& TGBH~\cite{di2020grand} & fMRI & 2020, bioRxiv & \multicolumn{1}{c|}{1} & The Grand Budapest Hot & 25 & & V2V \\ \cline{2-9} 

& CLSR~\cite{tang2023semantic} & fMRI & 2023, Nature Neuroscience & \multicolumn{1}{c|}{82} & {\makecell{Silent video clips \&\\ stories from \\The Moth Radio Hour and\\ Modern Love}} & 3 & AC, Broca, sPMv & V2T, S2T \\ 

\hline

\end{tabular}
}

\footnotesize{

$1$ In the table, the image datasets are sourced from: MNIST~\cite{deng2012mnist}, Scene Understanding(SUN)~\cite{zhou2017places}}, COCO~\cite{lin2014microsoft}, ImageNet~\cite{deng2009imagenet}, Commercial Digital Library~\cite{magazine2001commercial}, Berkeley Segmentation Dataset~\cite{arbelaez2010contour}, CelebA~\cite{sun2014deep}, VGGSound~\cite{2020Vggsound}, TIMIT~\cite{garofolo1993darpa}\\

\label{tabel1_dataset}
\end{table*}

%% file: tabels/model.tex
\begin{table*}[t]
\centering
\caption{Overview of Models for fMRI-Based Brain Decoding Tasks}
\renewcommand\arraystretch{1.3}
\begin{tabular}{|c|c|c|c|c|c|c|c|}
\hline
\thickhline
&   Methods  & Venue & Category & Architecture & Generate & Dataset & Task \\ \hline
\multirow{4}{*}{\rotatebox{90}{2019}} 
                      & IRM~\cite{shen2019end}& FCN & End-to-End & fMRI Features & GAN & DIR & I2I \\
                      & SSNIR~\cite{beliy2019voxels} & NeurIPS & Encoder-A-Based & DeCNN+fMRI & CAE & GOD & I2I \\ 
                      & DIR~\cite{shen2019deep} & PLoS CB & Pre-Trained & SLR+VGG & GAN & GOD & I2I \\
                      & BrainViVAE~\cite{han2019variational} & NeuroImage & Pre-Trained & LR+Latent Vectors & VAE & GOD & V2V \\ 
                      \hline
\multirow{2}{*}{\rotatebox{90}{2020}} 
                      & BrainSSG~\cite{fang2020reconstructing} & NeurIPS & Hybrid & FCN+DNN+I2I-GAN~\cite{isola2017image} & GAN & GOD & I2I \\
                      & BrainBBG~\cite{mozafari2020reconstructing} & IJCNN & Hybrid & Ridge+Latent Vectors & GAN & GOD & I2I \\ \hline
\multirow{3}{*}{\rotatebox{90}{2021}} 
                      & SSDR~\cite{gaziv2021more} & ArXiv & Encoder-A-Based & DeCNN+fMRI & CAE & GOD & I2I \\
                      & BrainDVG~\cite{ren2021reconstructing} & NeuroImage & Encoder-A-Based & VAE+GAN & VAE\&GAN & GOD & I2I \\ 
                      & GIC-PTL~\cite{huang2021neural} & NeuralNetworks & Encoder-A-Based & CNN & AR & OCD & I2T \\ 
                      \hline
\multirow{10}{*}{\rotatebox{90}{2022}}
                      & GIC-CT~\cite{zhang2022cnn} & CMPB & End-to-End & CNN+fMRI Features & Transformer & OCD & I2T \\
                      & f-CVGAN~\cite{wang2022reconstructing} & CC & End-to-End & FCN+fMRI Features & GAN & DNV & V2V \\
                      & Brain2Pix~\cite{le2022brain2pix} & FN & End-to-End & fMRI Features & GAN & STNS & V2V \\
                      & SSNIR-SC~\cite{gaziv2022self} & NeuroImage & Encoder-A-Based & DeCNN+fMRI & CAE & GOD & I2I \\
                      & SSRNM~\cite{kupershmidt2022penny} & ArXiv & Encoder-A-Based & DeCNN+fMRI & CAE & DNV & V2V \\
                      & SBD~\cite{ferrante2022semantic} & ArXiv & Hybrid & ResNet+Ridge & Diffusion & GOD & I2I \\
                      & BrainSCN~\cite{gu2022decoding} & ArXiv & Hybrid & SCN+ICGAN & GAN & NSD & I2I \\
                      & MindReader~\cite{lin2022mind} & NeurIPS & Pre-Trained & CLIP+StyleGAN2~\cite{karras2020analyzing} & GAN & NSD & I2I \\
                      & BrainICG~\cite{ozcelik2022reconstruction} & IJCNN & Hybrid & Ridge+SwAV & GAN & GOD & I2I \\
                      
                      \hline
\multirow{13}{*}{\rotatebox{90}{2023}}
                      & Mind-Vis~\cite{chen2023seeing} & CVPR & Hybrid & MAE+fMRI Features & Diffusion & GOD\&BLOD & I2I \\
                      & LEA~\cite{qian2023joint} & ArXiv & Pre-Trained & MAE+CLIP & AR & GOD\&BLOD & I2I \\ 
                      & UniCoRN~\cite{xi2023unicorn} & ArXiv & Hybrid & CAE+BART & AR & Narratives & S2T \\
                      & BrainCLIP~\cite{liu2023brainclip} & ArXiv & Pre-Trained & VAE+CLIP & Diffusion & NSD & I2I \\
                      & MindDiffuser~\cite{lu2023minddiffuser} & MM & Pre-Trained & Ridge+CLIP & Diffusion & NSD & I2I \\
                      & UniBrain~\cite{mai2023unibrain} & ArXiv & Pre-Trained & Ridge+CLIP & Diffusion & NSD & I2I \\
                      & DBDM~\cite{meng2023dual} & Bioengineering & Pre-Trained & MLP+CLIP & Diffusion & GOD & I2I \\
                      & BrainHVAE~\cite{miliotou2023generative} & ICML & Hybrid & FCN+VGG & VAE & GOD & I2I \\
                      & BSR~\cite{park2023sound} & ArXiv & Hybrid & Ridge+DNN+Transformer & AR & BSR & S2S \\
                      & Brain2Music~\cite{denk2023brain2music} & ArXiv & Hybrid & Ridge+MusicLM~\cite{agostinelli2023musiclm} & AR & MusicGenre & S2S \\
                      & DreamCatcher~\cite{chatterjee2023dreamcatcher} & ArXiv & LLM & FCN+GPT+LSTM & AR & NSD & I2T \\
                      & VDVAE~\cite{ozcelik2023natural} & ScientificReports & Pre-Trained & CLIP & Diffusion & NSD & I2I \\
                      & SDRecon~\cite{takagi2023high} & CVPR & Encoder-A-Based & CNN & LDM~\cite{rombach2022high} & NSD & I2I \\
                      \hline
\multirow{17}{*}{\rotatebox{90}{2024}}
                      & CMVDM~\cite{zeng2024controllable} & AAAI & Pre-Trained & MAE+ControlNet & Diffusion & GOD\&BLOD & I2I \\
                      & CAD~\cite{sun2024contrast} & NeurIPS & Hybrid & MAE+Attention & Diffusion & GOD\&BLOD & I2I \\ 
                      & MindVideo~\cite{chen2024cinematic} & NeurIPS & Pre-Trained & MAE+fMRI+CLIP & Diffusion & DNV & V2V \\
                      & MindEye~\cite{scotti2024reconstructing} & NeurIPS & Pre-Trained & MLP+CLIP & Diffusion & NSD & I2I \\
                      & Dream~\cite{xia2024dream} & WACV & Pre-Trained & CNN+MLP+CLIP & Diffusion & NSD & I2I \\
                      & BrainHSG~\cite{meng2024semantics} & TNSRE & Encoder-A-Based & DNN+VGG & GAN & GOD & I2I \\
                      & MusicCLAP~\cite{ferrante2024r} & ArXiv & Pre-Trained & CLAP+SWINTransformer & Diffusion & GTZan & S2S \\
                      & MindEye2~\cite{scotti2024mindeye2} & ArXiv & Pre-Trained & MLP+CLIP & Diffusion & NSD & I2I \\
                      & MindBridge~\cite{wang2024mindbridge} & CVPR & Pre-Trained & MLP+CLIP & VD~\cite{xu2023versatile} & NSD & I2I \\
                      & NeuralDiffuser~\cite{li2024neuraldiffuser} & ArXiv & Pre-Trained & MLP+CLIP & LDM~\cite{rombach2022high} & NSD & I2I \\
                      & BrainDiffuser~\cite{ferrante2024through} & ImagingNeuroscience & Pre-Trained & CLIP & Diffusion & NSD & I2I \\
                      & Psychometry~\cite{quan2024psychometry} & CVPR & Pre-Trained & Transformer+CLIP & Diffusion & NSD & I2I \\
                      & Cortex2Detail~\cite{gu2022decoding} & ArXiv & Encoder-A-Based & SwAV & Diffusion & NSD & I2I \\
                      & NeuroPictor~\cite{huo2025neuropictor} & ECCV & Pre-Trained & CLIP+Transformer & Diffusion & NSD & I2I \\
                      & UMBRAE~\cite{xia2025umbrae} & ECCV & LLM & LLM+Transformer & LDM~\cite{rombach2022high} & NSD & I2I \\
                      & BIR~\cite{ren2024brain} & PatternRecognition & Encoder-A-Based & StyleGAN~\cite{karras2020analyzing} & GAN & CelebrityFace & I2I \\
                      & DSPD-SNN~\cite{xu2024robust} & TNNLS & Encoder-A-Based & SNN & Diffusion & VER\&TGBH & V2V \\
                      \hline
\end{tabular}

\footnotesize{$1$ Due to space limitations, some venue names in the table are abbreviated. The abbreviations and their corresponding full journal names are as follows: FCN: Frontiers in Computational Neuroscience, PLoS CB: PLoS Computational Biology, CMPB: Computer Methods and Programs in Biomedicine, CC: Cerebral Cortex, FN: Frontiers in Neuroscience.}\\
\footnotesize{$2$ The model classification names are also abbreviated in the table due to length constraints, corresponding as follows: Encoder-A-Based: Encoder-Alignment-Based, Pre-Trained: Large-Scale Pre-Trained Generation, LLM: Large Language Model-Centric Models, Hybrid: Hybrid Models}

\label{tab:models}
\end{table*}

%% file: tabels/results.tex
\begin{table*}[t!]
\caption{Quantitative Results of I2I Tasks in fMRI-based Brain Decoding Tasks}
\tabcolsep 4pt
\renewcommand\arraystretch{1.1}
\resizebox{1.0\linewidth}{!}{
\begin{tabular}{>{\raggedleft\arraybackslash}p{2.7cm}>{\raggedright\arraybackslash}p{1.0cm}||c|cccc|cccc|c}
\hline
\thickhline
& &  & \multicolumn{4}{c|}{Low-Level} & \multicolumn{4}{c|}{High-Level} & \\ \cline{4-11}
\multirow{-2}{*}{Baseline} & &
  \multicolumn{1}{c|}{\multirow{-2}{*}{Subj}} &
  PixC $\uparrow$ & SSIM $\uparrow$ & Alex(2) $\uparrow$ & Alex(5) $\uparrow$ &
  Incep $\uparrow$ &  CLIP $\uparrow$ & EffNet-B $\downarrow$ &
  \multicolumn{1}{c|}{SwAV $\downarrow$} &
  \multirow{-3}{*}{Note} \\ \hline
\multicolumn{12}{c}{\cellcolor[HTML]{EFEFEF}NSD}               \\ \hline

MindReader~\cite{lin2022mind}\!&\!\pub{NeurIPS2022} & Avg.   &   -  &   -  &   -  &   -  & 78.2 &   -  &   -  &   -  &  \\ \hline

MindDiffuser~\cite{lu2023minddiffuser}\!&\!\pub{MM2023} & Avg.   &   -  & .354 &   -  &   -  &   -  & 76.5 &   -  &   -  \\ \hline

BrainCaptioning~\cite{ferrante2023brain}\!&\!\pub{arXiv2023}  & Avg.   & .353 & .327 & 89.0 & 97.0 & 84.0 & 90.0 &   -  &   -  &  Subj 12 \\ \hline

BrainCLIP~\cite{liu2023brainclip}\!&\!\pub{arXiv2023} 
 & Avg.   &   -  &   -  &   -  &   -  & 86.7 & 94.8 &   -  &   -  &  \\ \hline   

VDVAE~\cite{ozcelik2023natural}\!&\!\pub{Sci.R2023} & Avg.   & .254 & .356 & 94.2 & 96.2 & 87.2 & 91.5 & .775 & .423 &  \\ \hline

& & 1      & .185 & .288 & 91.7 & 95.0 & 84.7 & 88.3 & .776 & .418 & \\
\multirow{-2}{*}{SecondSight~\cite{kneeland2023second}}\!&\!\multirow{-2}{*}{\pub{arXiv2023}} & Avg.   & .156 & .285 & 88.4 & 93.5 & 82.0 & 87.0 & .792 & .435 & \\ \hline

SDRecon~\cite{takagi2023high}\!&\!\pub{CVPR2023} & Avg.   &   -  &   -  & 83.0 & 83.0 & 76.0 & 77.0 &   -  &   -  &  \\ \hline

 & & 1      & \textbf{.390} & .337 & 97.4 & 98.7 & 94.5 & 94.6 & .630 & .358 & \\
\multirow{-2}{*}{MindEye~\cite{scotti2024reconstructing}}\!&\!\multirow{-2}{*}{\pub{CVPR2024}}  & Avg.   & .309 & .323 & 94.7 & 97.8 & 93.8 & 94.1 & .645 & .367 & \multirow{-2}{*}{}       \\ \hline

& & 1      & .376 & \textbf{.440} & 97.5 & 99.1 & 95.4 & 92.6 & \textbf{.612} & .341 & \\
\multirow{-2}{*}{MindEye2~\cite{scotti2024mindeye2}}\!&\!\multirow{-2}{*}{\pub{ArXiv2024}} & Avg.   & \textbf{.322} & \textbf{.431} & 96.1 & \textbf{98.6} & \textbf{95.4} & 93.0 & \textbf{.619} & \textbf{.344} & \multirow{-2}{*}{}       \\ \hline
 
& & 1      &   -  &   -  &   -  &   -  &   -  &   -  &   -  &   -  & \\
& & Avg.   & .148 & .259 & 86.9 & 95.3 & 92.2 & 94.3 & .713 & .413 & \multirow{-2}{*}{Single} \\ \cline{3-12} 
 & & 1      & .157 & .275 & 88.1 & 95.5 & 90.0 & 93.9 & .747 & .436 & \\
\multirow{-4}{*}{MindBridge~\cite{wang2024mindbridge}}\!&\!\multirow{-4}{*}{\pub{CVPR2024}} & Avg.   & .151 & .263 & 87.7 & 95.5 & 92.4 & 94.7 & .712 & .418 & \multirow{-2}{*}{Bridge} \\ \hline

& & 1      & .277 & .385 & \textbf{98.8} & 99.3 & \textbf{96.2} & 94.5 & .619 & \textbf{.334} & \\
\multirow{-2}{*}{NeuroPictor~\cite{huo2025neuropictor}}\!&\!\multirow{-2}{*}{\pub{ECCV2024}} & Avg.   & .229 & .375 & \textbf{96.5} & 98.4 & 94.5 & 93.3 & .639 & .350 & \multirow{-2}{*}{}       \\ \hline

& & 1      & .288 & .338 & 95.0 & 97.5 & 94.8 & 95.2 & .638 & .413 & \\
\multirow{-2}{*}{DREAM~\cite{xia2024dream}}\!&\!\multirow{-2}{*}{\pub{WACV2024}} & Avg.   & .274 & .328 & 93.9 & 96.7 & 93.7 & 94.1 & .645 & .418 & \multirow{-2}{*}{}       \\ \hline

& & 1      & .168 & .282 & 66.3 & 77.5 & 64.8 & 76.0 &   -  &   -  & \\
\multirow{-2}{*}{\makecell{BrainDiffuser~\cite{ferrante2024through}}}\!&\!\multirow{-2}{*}{\pub{IN2024}} & Avg.   & .163 & .278 & 64.6 & 76.5 & 64.1 & 75.6 &   - &    -  & \multirow{-2}{*}{Subj 135}\\ \hline

Psychometry~\cite{quan2024psychometry}\!&\!\pub{CVPR2024} & Avg.   & .297 & .340 & 96.4 & \textbf{98.6} & 95.8 & \textbf{96.8} & .628 & .345 &  \\ \hline   

Cortex2Detail~\cite{gu2022decoding}\!&\!\pub{PMLR2024} & Avg.   & .150 & .325 &   -  &   -  &   -  &   -  & .862 & .465 &  \\ \hline

UMBRAE~\cite{xia2025umbrae}\!&\!\pub{ECCV2024}
 & Avg.   & .283 & .341 & 95.5 & 97.0 & 91.7 & 93.5 & .700 & .393 &  \\ \hline

MindSemantix~\cite{ren2024mindsemantix}\!&\!\pub{ArXiv2024}
 & Avg.   & .299 & .333 & 93.1 & 96.8 & 93.8 & 94.3 & .686 & .392 &  \\ \hline

&  & 1 & .206 & .382 & - & - & 85.9 & - & - & .431 & \\
\multirow{-2}{*}{TROI~\cite{wang2025troi}}\!&\!\multirow{-2}{*}{\pub{ArXiv2025}} & Avg.   & .177 & .380 & - & - & 83.2 & - & - & .454 & \multirow{-2}{*}{}       \\ \hline

BrainGuard~\cite{tian2025brainguard}\!&\!\pub{AAAI2025}
 & Avg.   & .313 & .330 & 94.7 & 97.8 & 96.1 & 96.4 & .624 & .353 &  \\ \hline
 
\multicolumn{12}{c}{\cellcolor[HTML]{EFEFEF}GOD}          \\ \hline

IC-GAN~\cite{ozcelik2022reconstruction}\!&\!\pub{IJCNN2022}
& Avg.   & .672 & .491 &   -  &   -  & \textbf{74.2} & 33.0 &   -  &   -  &  \\ \hline

& & 1      & \textbf{.372} & \textbf{.414} &   -  &   -  &   -  &   -  &   -  &   -  & \\
\multirow{-2}{*}{VQ-fMRI~\cite{chen2023rethinking}}\!&\!\pub{ICML2023} & Avg.   & \textbf{.371} & .400 &   -  &   -  &   -  &   -  &   -  &   -  & \multirow{-2}{*}{}\\ \hline 

& & 1      & .084 &   -  &   -  &   -  &   -  &   -  &   -  &   -  & \\
\multirow{-2}{*}{LEA~\cite{qian2023joint}}\!&\!\multirow{-2}{*}{\pub{arXiv2023}} & Avg.   & .158 &   -  &   -  &   -  &   -  &   -  &   -  &   -  & \multirow{-2}{*}{}\\ \hline 

BrainCLIP~\cite{liu2023brainclip}\!&\!\pub{arXiv2023} 
 & Avg.   &   -  &   -  &   -  &   -  & 65.1 & \textbf{84.2} &   -  &   -  &  \\ \hline 

CMVDM~\cite{zeng2024controllable}\!&\!\pub{AAAI2024} & Avg.   &   -  & \textbf{.631} &   -  &   -  &   -  &   -  &   -  &   -  &  \\ \hline

\multicolumn{12}{c}{\cellcolor[HTML]{EFEFEF}BOLD5000}          \\ \hline
& & 1      & .343 &   -  &   -  &   -  &   -  &   -  &   -  &   -  & \\
\multirow{-2}{*}{LEA~\cite{qian2023joint}}\!&\!\pub{arXiv2023}
 & Avg.   & .281 &   -  &   -  &   -  &   -  &   -  &   -  &   -  & \multirow{-2}{*}{}\\ \hline 

 CMVDM~\cite{zeng2024controllable}\!&\!\pub{AAAI2024} & Avg.   &   -  & .534 &   -  &   -  &   -  &   -  &   -  &   -  &  \\ \hline
\end{tabular}}
\label{tab:results}
\end{table*}

%% file: tabels/results_video.tex
\begin{table}[]
\caption{Quantitative Results of V2V Tasks in fMRI-based Brain Decoding Tasks}
\renewcommand\arraystretch{1.3}
\begin{tabular}{p{3.5cm}p{0.5cm}p{0.5cm}p{0.7cm}p{0.6cm}p{0.5cm}}
\hline
\multicolumn{1}{c|}{\multirow{3}{*}{Baseline}} & \multicolumn{5}{c}{Result}                                      \\ \cline{2-6} 
\multicolumn{1}{c|}{}                          & \multicolumn{3}{c|}{Low-Level} & \multicolumn{2}{c}{High-Level} \\ \cline{2-6} 
\multicolumn{1}{c|}{}                                & SSIM↑     & MSE↓    & PSNR↑    & Rank↓          & ACC↑          \\ \hline
DNV\pub{TIP2004}\cite{wang2004image}                 & .160      & .090    &  -       & 34.2           & -             \\ 
f-CVGAN\pub{CC2022}\cite{wang2022reconstructing}     & .094      & .118    & 11.4     & -              & -             \\                                      
SSRNM\pub{ArXiv2022}\cite{kupershmidt2022penny}      & .102      & .136    & -        & 7.18           & -             \\
Mind-Video\pub{NeurIPS2024}\cite{chen2024cinematic}  & .171      & -       & -        & -              & .202          \\   
MindGrapher\pub{2024}\cite{quanmindgrapher}  & .239      & -       & -        & -              & -          \\ \hline 
\end{tabular}

\footnotesize{$1$ Although there are some V2V studies, early works were more inclined toward qualitative analysis rather than quantitative evaluation, and therefore, their results are not presented here.}\\
\label{tab:resultsvideos}
\end{table}